%% file: main.tex
\pdfoutput=1
\documentclass{article}
\usepackage{colm2024_conference}
\usepackage{microtype}
\usepackage{hyperref}
\usepackage{url}
\usepackage{xurl}
\usepackage{booktabs}
\usepackage{CJKutf8}
\usepackage{graphicx}
\usepackage{multirow}
\usepackage{array}
\usepackage{amsmath}
\usepackage{enumitem}
\usepackage[utf8]{inputenc}
\usepackage{caption}
\usepackage{subcaption}
\usepackage[table]{xcolor}
\usepackage{tabularx}
\usepackage{makecell}
\usepackage{arydshln}
\usepackage{flafter}

\usepackage[section]{placeins}
\makeatletter
\AtBeginDocument{\expandafter\renewcommand\expandafter\subsection\expandafter{\expandafter\@fb@secFB\subsection}}
\makeatother
\usepackage{textcomp}

\newcommand{\linkicon}[2]{%
  \raisebox{-0.2em}{\includegraphics[height=1.1em]{#1}}\quad
  \href{#2}{\texttt{#2}}%
}

\title{Hy-MT3: More Accurate, Robust and Practical\\Multilingual Translation Models in the Wild}

\author{
	\bf \large Tencent Hunyuan Team
}

\begin{document}
\begin{CJK*}{UTF8}{gbsn}
\maketitle

\begin{abstract}
Hy-MT3 is the third generation of the Hunyuan multilingual translation model family, including \textbf{Hy-MT3-1.8B}, \textbf{Hy-MT3-7B} and \textbf{Hy-MT3-30B-A3B (MoE)}. Compared with Hy-MT2, Hy-MT3 translates \textbf{more accurately, more completely and more faithfully to user constraints}: on real-world scenario data, the share of samples with mistranslations is reduced by \textbf{43\%}, omissions or additions by \textbf{74\%}, and instruction-following errors by \textbf{40\%}, while language coverage is expanded from \textbf{38 to 102} languages.
With only 3B activated parameters, \textbf{Hy-MT3-30B-A3B achieves the best overall performance among all evaluated models in non-thinking mode}, including flagship models such as Qwen3.8-Max, Kimi-K3.1-Pro and DeepSeek-V4.1-Flash, and approaches the strongest closed-source thinking models such as GPT-6.1-Sol and Gemini-4-Argon, while surpassing models that consume \textbf{more than 30$\times$} its inference compute.
Beyond general translation, Hy-MT3 is optimized for two industrial scenarios. For \textbf{game localization}, built with high-quality game-domain data and a professional evaluation system from the Honor of Kings team, Hy-MT3-30B-A3B matches or outperforms strong closed-source models such as Gemini 3.1 Pro on all core metrics, and can be plugged into game development pipelines through the open-source toolchain TiMiHOKGameTranslator. For \textbf{image translation}, Hy-MT3 improves context understanding and error correction to better tolerate OCR errors, achieving the best result among models of comparable size.
Finally, we open-source \textbf{TransRubric}, a rubric-based automatic evaluation suite for translation, which outperforms other evaluation schemes on the WMT26 metrics shared task test set and is more stable with respect to human evaluation.

\vspace{0.5em}
\linkicon{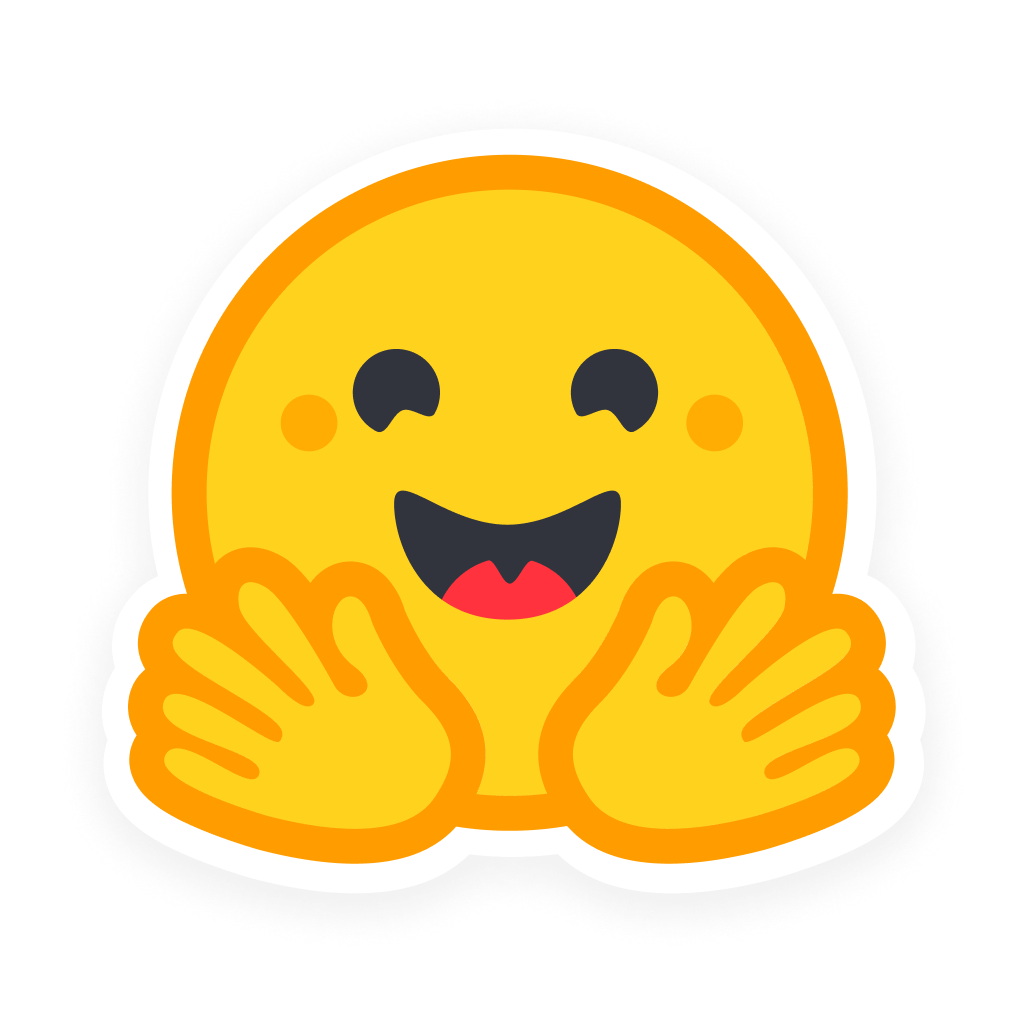}{https://huggingface.co/collections/tencent/hy-mt3}

\linkicon{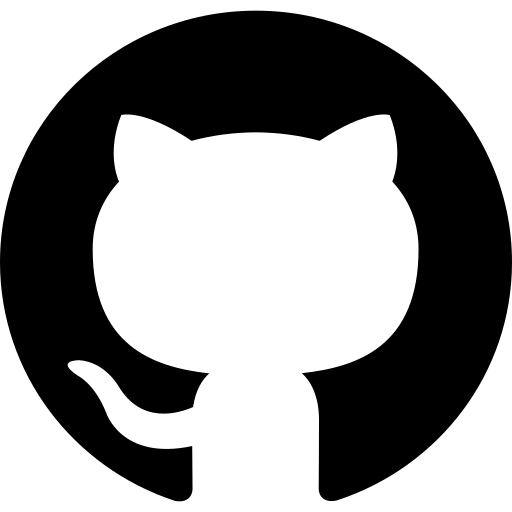}{https://github.com/Tencent-Hunyuan/Hy-MT3}
\end{abstract}

\begin{center}
\includegraphics[width=0.9\linewidth]{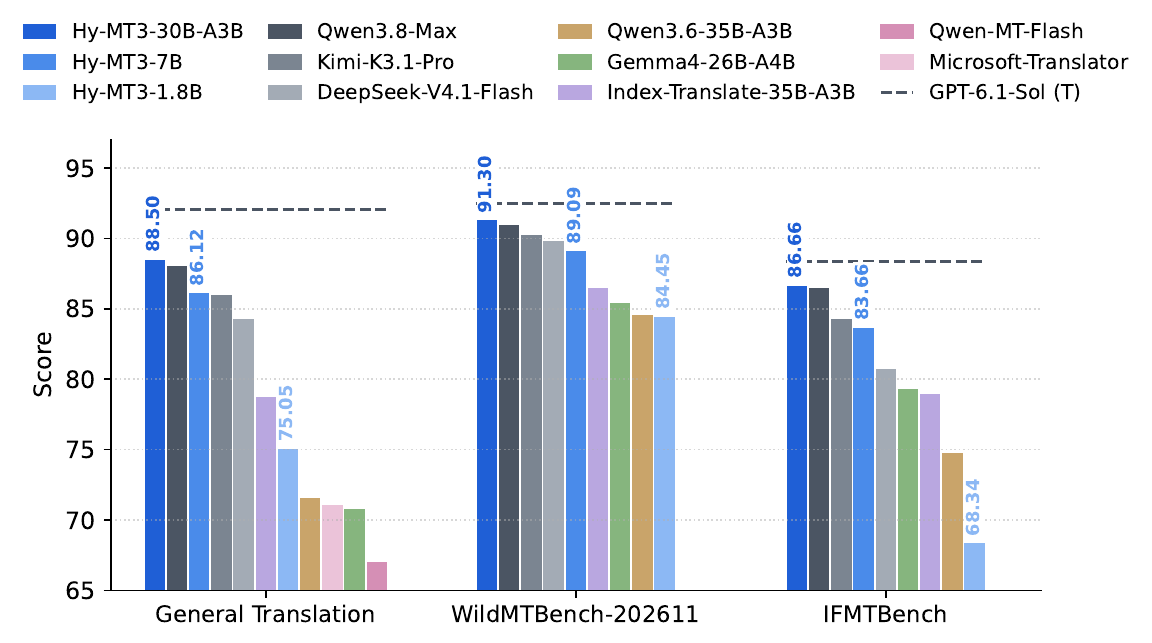}
\captionof{figure}{Overall performance of Hy-MT3 and representative baselines. General Translation is the average over FLORES-200, FLORES-200 low-resource, WMT26 and Mandarin$\Leftrightarrow$Minority translation (Table~\ref{tab:general_translation}); WildMTBench-202611 is built from the latest real-world scenario data; IFMTBench measures translation instruction following. General Translation and WildMTBench-202611 are scored by TransRubric. The dashed lines mark GPT-6.1-Sol (T) as a thinking-mode reference. (T) denotes thinking mode; all other models, including Hy-MT3, are run in non-thinking mode.}
\label{fig:overall}
\end{center}

\section{From Hy-MT2 to Hy-MT3}

Building on Hy-MT1.5~\citep{hy-mt1.5}, Hy-MT2~\citep{hy-mt2} introduced a family of fast-thinking translation models (1.8B, 7B and 30B-A3B) for real-world translation that follow translation instructions in multiple languages. As Hy-MT2 was adopted by the open-source community and in more production scenarios, several limitations became apparent.

First, \textbf{faithfulness remains the main source of errors in real-world translation}. A fine-grained analysis of Hy-MT2-30B-A3B on 5{,}000 samples of real-world scenario data (Table~\ref{tab:wild_breakdown}) shows that 56.0\% of its outputs contain at least one accuracy issue and 21.2\% contain omissions or additions. Such errors are rarely captured by surface-level fluency, but they directly determine whether a translation can be used without post-editing.

Second, \textbf{language coverage is insufficient for long-tail demand}. Hy-MT2 covers 38 languages and performs poorly on lower-resource languages outside this set: on 64 additional lower-resource languages paired with Chinese or English, Hy-MT2-30B-A3B obtains a TransRubric score of only 63.89 (ZH$\Leftrightarrow$XX) and 67.75 (EN$\Leftrightarrow$XX).

Third, \textbf{vertical scenarios impose requirements that general benchmarks do not measure}. Game localization requires strict preservation of placeholders and rich-text tags as well as consistent use of in-game terminology, and a single off-target or broken tag can break the game client. Image translation takes OCR output as input, which often contains misrecognized characters, missing text or wrong segmentation that the translation model must tolerate and correct from context.

Finally, \textbf{reliable automatic evaluation becomes the bottleneck}. As strong models converge on standard benchmarks, holistic scores and neural metrics~\citep{xcomet,cometkiwi} struggle to distinguish them or to explain where they fail. We need an evaluation method that is fine-grained, interpretable and consistent with human judgments, both to measure progress and to guide model development.

To address these issues, we present Hy-MT3. Our main contributions are as follows:
\begin{itemize}[leftmargin=1.5em]
    \item \textbf{More accurate, complete and constraint-faithful translation.} Compared with Hy-MT2-30B-A3B, the share of samples with mistranslations, omissions/additions and instruction-following errors is reduced by 43\%, 74\% and 40\%, respectively, and language coverage is expanded from 38 to 102 languages (Appendix~\ref{app:languages}).
    \item \textbf{Outperforming much larger models with high efficiency.} Hy-MT3-30B-A3B achieves the best overall performance among all evaluated models in non-thinking mode, including Qwen3.8-Max, Kimi-K3.1-Pro and DeepSeek-V4.1-Flash, and approaches the strongest closed-source thinking models such as GPT-6.1-Sol and Gemini-4-Argon.
    \item \textbf{Game translation.} Together with the Honor of Kings team, we improve game translation with high-quality game-domain data and a professional evaluation system. Hy-MT3-30B-A3B matches or outperforms strong closed-source models such as Gemini 3.1 Pro on all core metrics, and an open-source toolchain allows it to be integrated into game development pipelines for high-quality, low-cost localization.
    \item \textbf{Image translation.} Hy-MT3 is optimized for OCR-cascaded image translation, with stronger context understanding and error correction to tolerate OCR errors, achieving the best result among models of comparable size.
    \item \textbf{TransRubric.} We open-source TransRubric, an automatic evaluation suite for translation, which outperforms other evaluation schemes on the WMT26 metrics shared task test set and is more stable with respect to human evaluation.
\end{itemize}

\section{TransRubric: Rubric-Based Automatic Evaluation for Translation}\label{sec:transrubric}

To evaluate translation quality in a fine-grained and interpretable way, we develop TransRubric, a rubric-based LLM-as-a-judge~\citep{zheng2023judging,gembda} evaluation suite, and use it as the primary metric throughout this report. Given a source text, a candidate translation, the translation instruction and optionally a reference translation, a TransRubric judge checks the candidate against a translation rubric, identifies each issue together with its error type (e.g., completeness, accuracy, terminology and instruction following) and severity level, in the spirit of MQM-style error annotation~\citep{14-MQM,21-wmt-mqm}, and outputs a pointwise score on a 0--100 scale. The issue annotations make the score interpretable and enable error analyses such as Table~\ref{tab:wild_breakdown}. TransRubric supports both reference-based evaluation (used for FLORES-200, WMT26, Mandarin$\Leftrightarrow$Minority and HardMTBench) and reference-free evaluation (used for WildMTBench-202611, which has no reference translations).

In this report, each TransRubric score is the arithmetic mean of two judges, DeepSeek-V4.1-Flash~\citep{deepseekv4.1} and Qwen3.5-397B-A17B~\citep{qwen3.5}. To make scores comparable across models, each benchmark score is computed with the fixed expected number of samples as the denominator, so empty outputs and unscored samples count as 0.

\textbf{Meta-evaluation on WMT26.} We validate TransRubric on Task 2 (segment-level quality score prediction) of the WMT26 shared task on automated translation quality evaluation~\citep{wmt26-mteval}, following its official meta-evaluation on the eight priority translation directions: screen-based Pairwise Difference Pearson (PDP) at the segment level, source-based Soft Pairwise Accuracy (SPA) at the system level, and their average (Avg Corr). In this experiment, TransRubric runs reference-free with Gemini 3.6 Flash as the underlying judge. As shown in Table~\ref{tab:transrubric_meta}, TransRubric achieves the highest PDP (0.744) and Avg Corr (0.832), ahead of the top-ranked official submission Gemini-3.6-Flash (0.736 / 0.830), which uses the same underlying model, as well as GPT-5.6 Sol (0.828) and the ensemble system bytepop\_pro10 (0.826). Its system-level SPA (0.920) is on par with the best system (0.924), and all LLM-based evaluators clearly outperform conventional metrics such as chrF, COMET22 and xCOMET.

\input{tables/transrubric_meta}

\begin{figure}[!htbp]
\centering
\includegraphics[width=\linewidth]{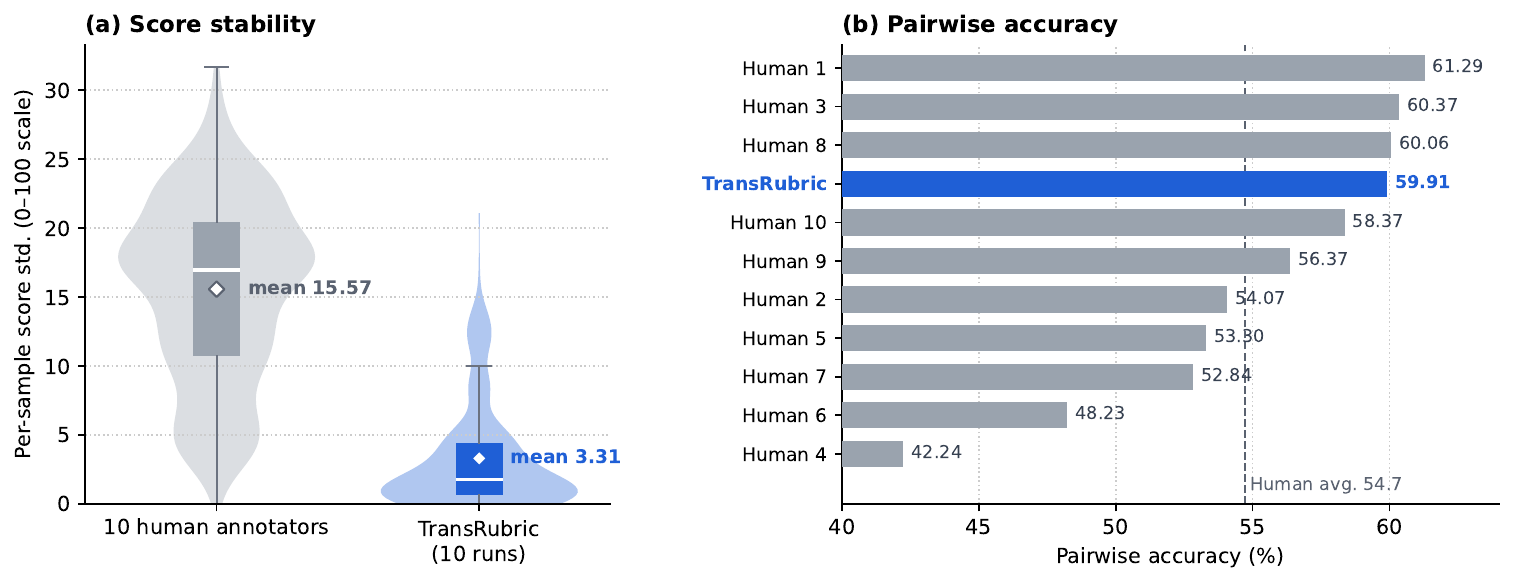}
\caption{TransRubric versus 10 human annotators on 651 Chinese$\Leftrightarrow$English translations. (a) Distribution of the per-sample standard deviation of the 10 scores, across annotators for humans and across 10 repeated runs for TransRubric; diamonds mark the means. (b) Pairwise accuracy of each rater; the dashed line marks the average of the 10 annotators.}
\label{fig:transrubric_human}
\end{figure}

\textbf{Comparison with human annotators.} We further compare TransRubric with 10 professional annotators majoring in English on 651 translations, i.e., 217 source texts with three candidate translations each, covering both Chinese$\rightarrow$English and English$\rightarrow$Chinese. Each annotator scores every translation once on a 0--100 scale, and TransRubric (with the DeepSeek-V4.1-Flash and Qwen3.5-397B-A17B judges) scores every translation 10 times. Figure~\ref{fig:transrubric_human} summarizes the results.
\begin{itemize}[leftmargin=1.5em]
    \item \textbf{Stability.} For each translation, we compute the standard deviation of its 10 scores. Across the 10 human annotators, the per-sample standard deviation averages 15.57 points, whereas across the 10 TransRubric runs it averages only 3.31 points (average variance 285.79 vs.\ 26.20). TransRubric scores are thus far more consistent than human judgments.
    \item \textbf{Pairwise accuracy.} For the three translations of each source text, we form all pairs (651 pairs in total) and check whether each rater prefers the same translation as the gold standard. The gold score of each translation is obtained by English translation experts, who revise the 10 annotators' scores and take the average of the revised scores. Since the gold standard is built on the annotators' own scores, this comparison may favor the human annotators. TransRubric reaches 59.91\%, ranking 4th among the 11 raters: it is higher than 7 of the 10 annotators and lower only than three of them (61.29\%, 60.37\% and 60.06\%), well above the human average of 54.7\%.
\end{itemize}

TransRubric is released together with Hy-MT3 at \url{https://github.com/Tencent-Hunyuan/Hy-MT3}.

\section{Experiments}

\subsection{Experimental Setup}

\textbf{Benchmarks.} We evaluate Hy-MT3 from three perspectives.
\begin{itemize}[leftmargin=1.5em]
    \item \textbf{General translation.} FLORES-200~\citep{flores-200} on 33 languages with all $33\times32=1056$ directions; FLORES-200 on 64 additional lower-resource languages paired with Chinese or English (256 directions); WMT26 GenMT~\citep{wmt26-genmt} on 20 language pairs after excluding English$\rightarrow$Ligurian, English$\rightarrow$Ladin and English$\rightarrow$Northern Sami; and Mandarin$\Leftrightarrow$Minority translation between Mandarin and Tibetan, Uyghur, Traditional Mongolian, Arabic-script Kazakh and Cantonese (10 directions).
    \item \textbf{Real-world translation.} WildMTBench-202611, the latest version of our in-the-wild benchmark, consists of 5{,}000 samples of real-world scenario data evenly distributed over 10 domains; HardMTBench~\citep{hardmtbench} is an open-source benchmark of knowledge-intensive, long-form Chinese--English cases in both directions\footnote{\url{https://github.com/jasonNLP/HardMTBench}}.
    \item \textbf{Translation instruction following.} IFMTBench~\citep{hy-mt2}\footnote{\url{https://github.com/Tencent-Hunyuan/Hy-MT2/tree/main/IFMTBench}} contains 7{,}344 samples covering industrial translation constraints such as code tags, inline code, layout preservation, style, glossary, context and structured data, split into single-constraint (Simple) and multi-constraint (Complex) subsets.
\end{itemize}
In addition, we report human evaluation on game translation (Section~\ref{sec:game}) and cascaded image translation on VISTA-Bench~\citep{lv2026vistabenchbenchmarkingmultilingualimage} (Section~\ref{sec:image}).

\textbf{Metrics.} We report TransRubric (Section~\ref{sec:transrubric}) as the primary metric, and GEMBA~\citep{gembda} with GPT-OSS-120B~\citep{gptoss} (medium reasoning) as a complementary LLM-based metric. IFMTBench is scored with its official rules and LLM judge.

\textbf{Baselines.} We compare Hy-MT3 with Hy-MT2 and three groups of baselines: (1) general models in non-thinking mode, including Kimi-K3.1-Pro~\citep{kimik3.1}, DeepSeek-V4.1-Flash~\citep{deepseekv4.1}, the Qwen3.8~\citep{qwen3.8,qwen3.8next} / Qwen3.6~\citep{qwen3.6} series and the Gemma4 series~\citep{gemma4}; (2) translation-specialized models and commercial translation APIs, including the Index-Translate series~\citep{indextranslate}, Tower-Plus-72B~\citep{towerplus}, Qwen-MT-Flash~\citep{qwenmt}, Microsoft Translator, Volcano Translator and iFlytek Translator; and (3) models in thinking mode, including GPT-6.1-Sol~\citep{gpt6.1sol}, Claude-Opus-5.5~\citep{claudeopus5.5}, Gemini-4-Argon~\citep{gemini4argon}, GLM-5.3~\citep{glm5,glm5.3} and the thinking variants of the open-weight models above. All Hy-MT3 models are run in non-thinking mode, so the most direct comparison is with groups (1) and (2).

\subsection{General Translation}

Table~\ref{tab:general_translation} reports results on general translation benchmarks.

\input{tables/general_translation}

\textbf{Hy-MT3-30B-A3B achieves the best overall result among all non-thinking systems.} Its average TransRubric / GEMBA scores reach 88.50 / 81.87, ahead of Qwen3.8-Max (88.02 / 81.08), Kimi-K3.1-Pro (86.01 / 81.47) and DeepSeek-V4.1-Flash (84.27 / 78.98). On the 33-language FLORES-200 set, Hy-MT3-30B-A3B obtains the best TransRubric scores among non-thinking systems on all three subsets (94.06 / 94.82 / 92.12 on ZH$\Leftrightarrow$XX, EN$\Leftrightarrow$XX and XX$\Leftrightarrow$XX).

\textbf{Language expansion brings large gains on lower-resource languages.} On the 64 additional lower-resource languages, Hy-MT3-30B-A3B improves the TransRubric score over Hy-MT2-30B-A3B from 63.89 to 91.52 (ZH$\Leftrightarrow$XX), the best among non-thinking systems, and from 67.75 to 92.74 (EN$\Leftrightarrow$XX), close to Qwen3.8-Max (92.99). The gains are even larger for smaller models: Hy-MT3-1.8B improves from 24.44 to 73.48 on ZH$\Leftrightarrow$XX.

\textbf{Hy-MT3 is strong on WMT26 and Mandarin$\Leftrightarrow$Minority translation.} On WMT26, Hy-MT3-30B-A3B improves over Hy-MT2-30B-A3B from 73.03 to 85.63 in TransRubric, close to the strongest non-thinking systems Qwen3.8-Max (86.88) and Kimi-K3.1-Pro (85.86). On Mandarin$\Leftrightarrow$Minority translation, it achieves 83.85 / 70.17, the best among non-thinking systems on both metrics.

\textbf{Smaller models remain highly competitive.} Hy-MT3-7B reaches an average of 86.12 / 80.39, surpassing DeepSeek-V4.1-Flash and all translation-specialized models, including Index-Translate-35B-A3B (78.75 / 79.24). Hy-MT3-1.8B improves the average TransRubric score of Hy-MT2-1.8B from 56.51 to 75.05, outperforming commercial APIs such as iFlytek Translator (73.26), Microsoft Translator (71.09) and Qwen-MT-Flash (67.00), as well as Index-Translate-2B (60.11).

\begin{figure}[!htbp]
\centering
\includegraphics[width=0.9\linewidth]{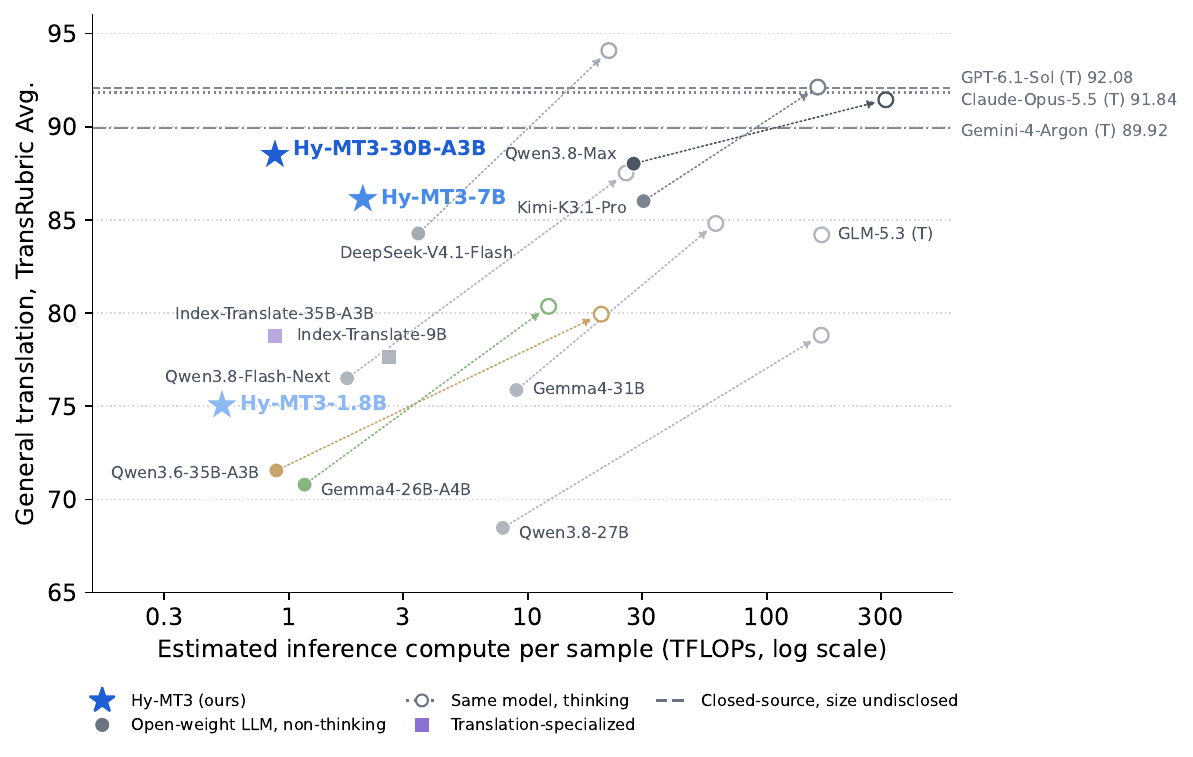}
\caption{General translation quality versus inference compute. The x-axis is the estimated inference compute per test sample, approximately $2\times$ activated parameters $\times$ generated tokens (including thinking tokens), in log scale. The y-axis is the average TransRubric score in Table~\ref{tab:general_translation}. Arrows connect the non-thinking and thinking modes of the same model; dashed horizontal lines denote closed-source models whose sizes are undisclosed.}
\label{fig:efficiency}
\end{figure}

\textbf{Efficiency.} Figure~\ref{fig:efficiency} plots translation quality against estimated inference compute. Hy-MT3 models lie on the upper-left frontier: Hy-MT3-30B-A3B requires less than 1 TFLOP per sample, yet outperforms models that consume more than 30$\times$ its compute, such as Qwen3.8-Max and Kimi-K3.1-Pro in non-thinking mode. Thinking mode improves the quality of general models, but typically requires several times to more than an order of magnitude more compute.

\subsection{Real-World, Hard-Case and Instruction-Following Translation}

Table~\ref{tab:wild_hard_if} reports results on WildMTBench-202611, HardMTBench and IFMTBench.

\input{tables/wild_hard_if}

\textbf{Real-world translation.} On WildMTBench-202611, Hy-MT3-30B-A3B achieves 91.30, the best among non-thinking systems, ahead of Qwen3.8-Max (90.94) and Kimi-K3.1-Pro (90.23). It also outperforms several thinking models, such as Gemini-4-Argon$^{\mathrm{T}}$ (90.30) and Qwen3.8-Flash-Next$^{\mathrm{T}}$ (90.40), and is close to GPT-6.1-Sol$^{\mathrm{T}}$ (92.50). Notably, Hy-MT3-1.8B (84.45) already surpasses the previous-generation Hy-MT2-30B-A3B (83.62).

\textbf{Hard, knowledge-intensive translation.} On HardMTBench, Hy-MT3-30B-A3B improves over Hy-MT2-30B-A3B from 81.38 to 88.93 in TransRubric and clearly outperforms models of comparable size, such as Index-Translate-35B-A3B (85.57), Qwen3.6-35B-A3B (84.26) and Gemma4-26B-A4B (83.01). However, a gap remains compared with much larger models such as Qwen3.8-Max (92.30) and Kimi-K3.1-Pro (92.04). This suggests that in knowledge-intensive scenarios, larger models benefit from their greater parameter capacity to store world and domain knowledge.

\textbf{Translation instruction following.} On IFMTBench, Hy-MT3-30B-A3B achieves 91.02 / 79.74 / 86.66 on Simple, Complex and Total, obtaining the best Simple and Total scores among non-thinking systems and surpassing thinking models such as Qwen3.8-Max$^{\mathrm{T}}$ (86.32), DeepSeek-V4.1-Flash$^{\mathrm{T}}$ (85.37) and Claude-Opus-5.5$^{\mathrm{T}}$ (82.82) in Total. On Complex instructions, it still trails Qwen3.8-Max (82.84) and Kimi-K3.1-Pro (82.21), indicating room for improvement under multi-constraint instructions. Hy-MT3-7B (83.66) also surpasses Hy-MT2-30B-A3B (83.35).

\subsection{Domain-Wise Analysis}

\begin{figure}[!htbp]
\centering
\includegraphics[width=0.85\linewidth]{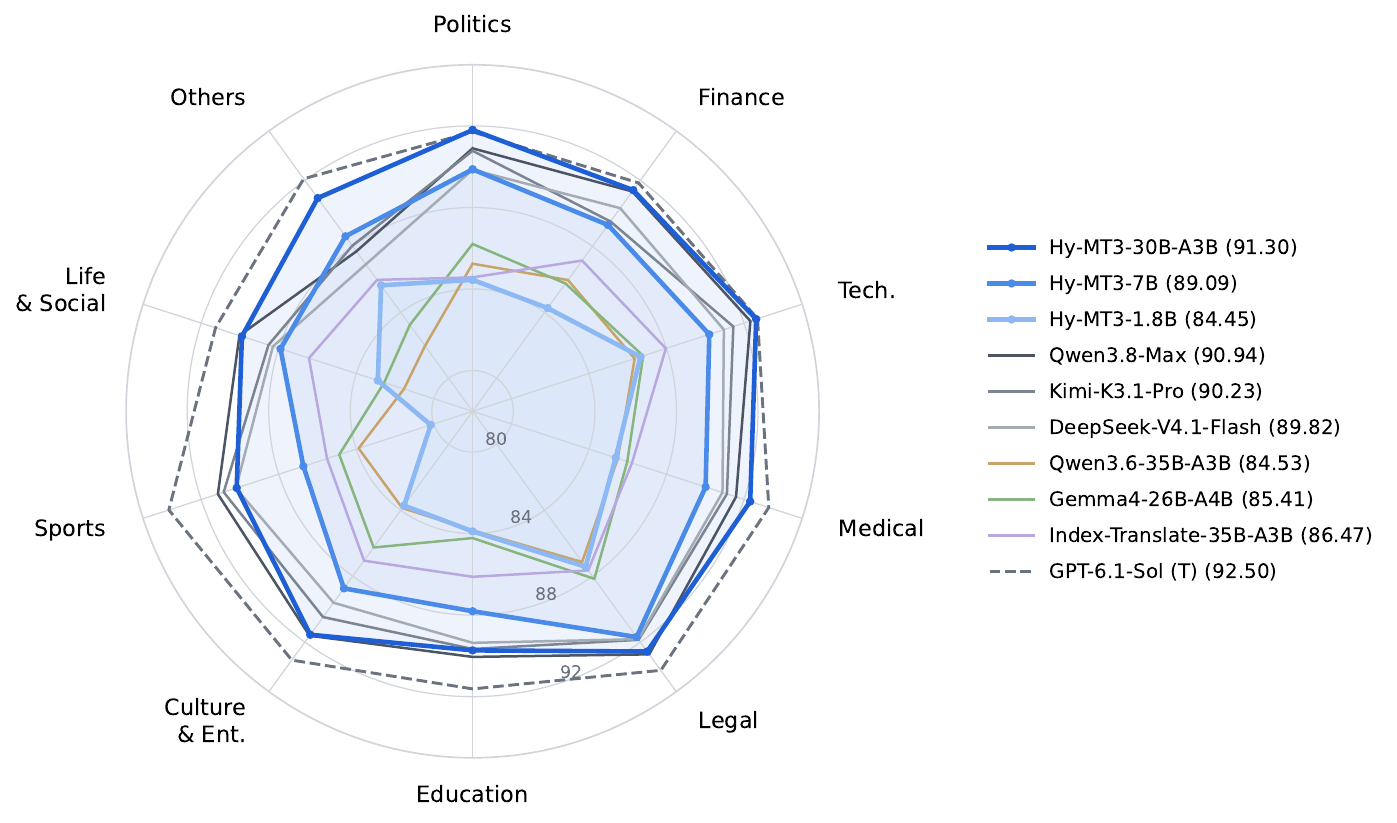}
\caption{Domain-wise TransRubric scores on WildMTBench-202611. Numbers in parentheses are the overall averages. (T) denotes thinking mode.}
\label{fig:wild_domain}
\end{figure}

We further break down WildMTBench-202611 into 10 domains (Table~\ref{tab:wild_domain} and Figure~\ref{fig:wild_domain}). Hy-MT3-30B-A3B achieves the best score among non-thinking systems in 5 of the 10 domains (politics, finance, technology, medical and others) and is within 0.35 points of the best in legal, education, culture \& entertainment and life \& social. The largest margin appears in the ``others'' domain, where Hy-MT3-30B-A3B leads Qwen3.8-Max by 3.23 points (90.93 vs.\ 87.70), indicating strong robustness on long-tail inputs. The largest remaining gap is in sports (90.17 vs.\ 91.14 for Qwen3.8-Max).

\input{tables/wild_domain}

\subsection{Improvements over Hy-MT2}

To understand where the improvements come from, we compare Hy-MT2-30B-A3B and Hy-MT3-30B-A3B on WildMTBench-202611 using the issue annotations produced by TransRubric (Table~\ref{tab:wild_breakdown}).

\input{tables/wild_breakdown}

\textbf{Score distribution.} Hy-MT3 raises the average score from 83.62 to 91.30 (+7.68). The share of samples scoring at least 90 increases from 46.1\% to 70.9\%, and the poor-translation rate (score below 60) drops from 7.4\% to 2.2\%. In a per-sample comparison, Hy-MT3 wins on 58.9\% of the samples and loses on only 15.5\%.

\begin{figure}[!htbp]
\centering
\includegraphics[width=0.85\linewidth]{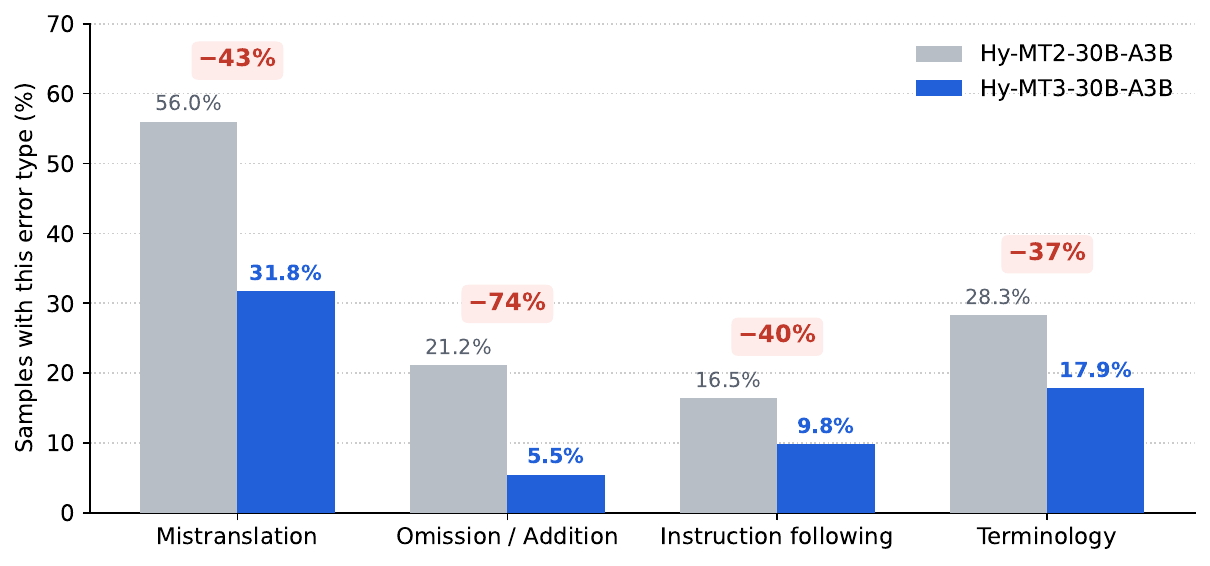}
\caption{Share of samples containing each error type on WildMTBench-202611, Hy-MT2-30B-A3B versus Hy-MT3-30B-A3B (Table~\ref{tab:wild_breakdown}b). Labels above each pair give the relative reduction.}
\label{fig:error_reduction}
\end{figure}

\textbf{Error types.} The improvements mainly come from faithfulness (Figure~\ref{fig:error_reduction}). The share of samples with omissions or additions drops from 21.2\% to 5.5\% ($-$74\%), mistranslations from 56.0\% to 31.8\% ($-$43\%), terminology issues from 28.3\% to 17.9\% ($-$37\%), and instruction-following issues from 16.5\% to 9.8\% ($-$40\%). The reduction in instruction-following issues mainly comes from Hy-MT3 no longer completing or extending the source text on its own, and from better preservation of formats and tags.

\subsection{Case Study}\label{sec:case_study}

Figure~\ref{fig:case_study} compares Hy-MT2 and Hy-MT3 on representative cases of the three error types where Hy-MT3 improves the most.

\textbf{Mistranslation.} Hy-MT2 misreads the person name ``夏雨雨'' as ``summer rains'' and over-interprets ``动容'' as ``moved to tears'', while Hy-MT3 correctly transliterates the name and keeps the original meaning. In the marketing copy, Hy-MT2 adds embellishments that are not in the source (``卓越功效'', ``完美''), whereas Hy-MT3 translates ``powerful pairing'' and ``designed to support'' faithfully.

\textbf{Omission.} Hy-MT2 drops key information such as ``Chippy Exhibition'' and ``硬核'', while Hy-MT3 preserves it (``对抗激烈的表演赛'', ``Hardcore shots'') together with the inline tags.

\textbf{Instruction following.} Hy-MT3 keeps the author list and journal name of a bibliographic reference in their original form, following the citation convention, while Hy-MT2 translates them. When asked for a colloquial translation, Hy-MT3 produces a more conversational rendering than the formal output of Hy-MT2.

\input{tables/case_study}

\subsection{Game Translation}\label{sec:game}

Game localization places strict requirements on output format and terminology: placeholders and rich-text tags must be preserved exactly, in-game terms must be translated consistently, and any off-target output can directly break the game client. Together with the Honor of Kings team, we collected high-quality game-domain data to make Hy-MT3-30B-A3B a leading model for game translation. We also jointly built an evaluation system covering multiple game genres. The evaluation set is stratified by localization characteristics such as text length, rich text and placeholders, and includes long texts with high terminology density from MOBA and open-world games, to comprehensively test how well a model handles game context, terminology and complex text structures. Under this system, we establish four game-specific evaluation criteria: off-target rate, placeholder errors, rich-text errors and terminology errors.

As shown in Table~\ref{tab:game}, Hy-MT3-30B-A3B achieves the lowest error rate on all four criteria (tied with Gemini 3.1 Pro on placeholder errors) and the highest total score in human evaluation, outperforming strong thinking models such as Gemini 3.1 Pro (88.7), GPT-5.5 (88.6) and DeepSeek-V4-Pro (88.3). In particular, its terminology error rate is 0.08\%, compared with 0.54\% for Gemini 3.1 Pro.

The Honor of Kings team has also open-sourced a game translation toolchain\footnote{\url{https://github.com/Tencent/TiMiHOKGameTranslator}} that supports multiple translation platforms, multiple file formats, pluggable workflows and various code management methods. It seamlessly embeds Hy-MT3-30B-A3B into the game development pipeline, connecting translation with downstream engineering. After integration, localization requires no additional manual effort, providing a reusable reference solution for game localization.

\input{tables/game}

\subsection{Image Translation}\label{sec:image}

In image translation, text in an image is typically first recognized by an OCR engine and then translated. OCR output often contains misrecognized characters, missing text or wrong segmentation, so the translation model must infer the intended content from context. We optimize Hy-MT3 for this scenario by strengthening its context understanding and error correction.

We evaluate on VISTA-Bench~\citep{lv2026vistabenchbenchmarkingmultilingualimage} in a cascaded setting with PaddleOCR~\citep{paddleocr3}. As shown in Table~\ref{tab:vista_bench}, Hy-MT3-30B-A3B achieves 45.73, the best among non-thinking and translation-specialized models, ahead of Gemma4-26B-A4B (45.46) and Index-Translate-35B-A3B (43.63), and also above the thinking model GLM-5.3$^{\mathrm{T}}$ (42.84).

\input{tables/vista}

\subsection{Quantization}\label{sec:quantization}

To support deployment under different resource budgets, we evaluate Hy-MT3 under several quantization settings: FP8, and the Q8, Q4\_K\_M and IQ2 formats of llama.cpp, ranging from 8-bit to 2-bit precision. Table~\ref{tab:quantization_results} compares them with the BF16 models.

\textbf{8-bit and 4-bit quantization is nearly lossless for the larger models.} For Hy-MT3-7B, FP8, Q8 and Q4\_K\_M stay within 0.6 points of BF16 on every benchmark; for Hy-MT3-30B-A3B, the gap is within 0.4 points, and some settings even score slightly higher than BF16.

\textbf{2-bit quantization remains usable.} With IQ2, the scores of Hy-MT3-7B and Hy-MT3-30B-A3B drop by at most 1.95 and 2.53 points, respectively, with the largest drops on low-resource FLORES-200, WMT26 and Mandarin$\Leftrightarrow$Minority translation. Hy-MT3-30B-A3B-IQ2 still outperforms the BF16 Hy-MT3-7B on all benchmarks except Mandarin$\Leftrightarrow$Minority translation.

\textbf{The smallest model is more sensitive to quantization.} For Hy-MT3-1.8B, Q8 stays within 0.6 points of BF16 except on IFMTBench ($-$1.60), whereas FP8 and Q4\_K\_M lower scores by up to 3.25 and 2.55 points, respectively, mainly on low-resource FLORES-200, Mandarin$\Leftrightarrow$Minority translation and IFMTBench. Its scores on WildMTBench-202611 and HardMTBench drop by less than 0.6 points under all settings.

\input{tables/quantization}

\section{Conclusion}

We present Hy-MT3, a family of multilingual translation models comprising Hy-MT3-1.8B, Hy-MT3-7B and Hy-MT3-30B-A3B. Compared with Hy-MT2, Hy-MT3 substantially reduces mistranslations, omissions/additions and instruction-following errors, and expands language coverage from 38 to 102 languages. Running in non-thinking mode with only 3B activated parameters, Hy-MT3-30B-A3B achieves the best overall performance among all evaluated non-thinking models on general translation, real-world translation and translation instruction following, and approaches the strongest closed-source thinking models at a fraction of their inference compute. Hy-MT3 further delivers leading results in game localization and OCR-cascaded image translation. Together with the models, we release TransRubric to support fine-grained and reliable evaluation of translation quality. We hope Hy-MT3 and TransRubric will facilitate both research and real-world applications of machine translation.

\section{Contributions}

\subsection{Core Contributors}
Mao Zheng, Zheng Li, Fangxu Liu, Tao Chen, Bingxin Qu, Bo Lv, Mingrui Sun

\subsection{Contributors}
Shiqi Huang, Hongyu Li, Hong Huang, Jinlong Song, Decheng Wu, Guanghua Yu, Shijia Wen, Jianwei Wu, Lizhi Fang, Sijia Wang, Zhongfu Zheng, Jing Zhao, Weiwei Xun, Yutong Li, Hengjian Long, Hanxin Liu, Dawei Wu, Jia Qing, Kaiyou He, Youning Liang

\appendix
\section{Supported Languages}\label{app:languages}

Table~\ref{tab:languages} lists the 102 languages supported by Hy-MT3, including the 38 languages already supported by Hy-MT2 and 64 newly supported languages.

\input{tables/languages}

\end{CJK*}

\input{main.bbl}
\end{document}

%% file: tables/transrubric_meta.tex
\begin{table}[!htbp]
\centering
\caption{Meta-evaluation on Task 2 (segment-level quality score prediction) of the WMT26 shared task on automated translation quality evaluation.}
\label{tab:transrubric_meta}
\scriptsize
\renewcommand{\arraystretch}{1.18}
\setlength{\tabcolsep}{5pt}
\begin{tabular}{llccc}
\toprule
\textbf{Auto-evaluator} & \textbf{Underlying model} & \textbf{PDP} & \textbf{SPA} & \textbf{Avg Corr} \\
\midrule
\textbf{TransRubric (ours)}$^{*}$ & Gemini 3.6 Flash & \colorbox{orange!20}{0.744} & 0.920 & \colorbox{orange!20}{0.832} \\
\hline
Gemini-3.6-Flash$^{*}$ & Gemini 3.6 Flash & 0.736 & \colorbox{orange!20}{0.924} & 0.830 \\
GPT-5.6 Sol - Medium Reasoning$^{*}$ & GPT-5.6 Sol & 0.735 & 0.920 & 0.828 \\
bytepop\_pro10 & EnsembleE2 & 0.733 & 0.918 & 0.826 \\
chrF & -- & 0.601 & 0.890 & 0.746 \\
COMET22 & XLM-RoBERTa-large & 0.610 & 0.862 & 0.736 \\
spBLEU & -- & 0.566 & 0.885 & 0.725 \\
gemba-poly$^{*}$ & Llama 3.3 70B & 0.510 & 0.884 & 0.697 \\
xCOMET XXL$^{\dagger}$ & xCOMET-XXL & 0.540 & 0.822 & 0.681 \\
BERTScore & XLM-RoBERTa-large & 0.496 & 0.858 & 0.677 \\
COMETKiwi22$^{*}$ & XLM-RoBERTa-large & 0.426 & 0.724 & 0.575 \\
\bottomrule
\end{tabular}
\vspace{1mm}
\begin{minipage}{0.98\linewidth}
\tiny
\textit{Notes.} Results of all systems other than TransRubric are taken from Table 8 of the WMT26 findings~\citep{wmt26-mteval}, which covers the eight priority translation directions. PDP is the screen-based Pairwise Difference Pearson at the segment level, SPA is the source-based Soft Pairwise Accuracy at the system level, and Avg Corr is their arithmetic mean; both are averaged equally across directions. $^{*}$ marks reference-free evaluators and $^{\dagger}$ fine-tuned systems. The best result in each column is highlighted with an \colorbox{orange!20}{orange background}.
\end{minipage}
\end{table}

%% file: tables/general_translation.tex
\begin{table*}[!htbp]
\centering
\caption{Performance comparison on general multilingual translation benchmarks.}
\label{tab:general_translation}
\scriptsize
\renewcommand{\arraystretch}{1.18}
\setlength{\tabcolsep}{4pt}
\resizebox{\textwidth}{!}{
\begin{tabular}{lcccccccc}
\toprule
\multirow{2}{*}{\textbf{Models}}
& \multicolumn{3}{c}{\textbf{FLORES-200} (33 languages, all pairs)}
& \multicolumn{2}{c}{\textbf{FLORES-200} (low-resource $\Leftrightarrow$ ZH/EN)}
& \multirow{2}{*}{\textbf{WMT26}}
& \multirow{2}{*}{\textbf{Mand.}$\Leftrightarrow$\textbf{Min.}}
& \multirow{2}{*}{\textbf{Avg.}} \\
\cmidrule(lr){2-4}\cmidrule(lr){5-6}
& ZH$\Leftrightarrow$XX & EN$\Leftrightarrow$XX & XX$\Leftrightarrow$XX
& ZH$\Leftrightarrow$XX & EN$\Leftrightarrow$XX & & & \\
\midrule
\textbf{Hy-MT3-30B-A3B} & \colorbox{orange!20}{94.06} / 90.72 & \colorbox{orange!20}{94.82} / 91.37 & \colorbox{orange!20}{92.12} / 89.14 & \colorbox{orange!20}{91.52} / 86.59 & 92.74 / 87.15 & 85.63 / 81.07 & \colorbox{orange!20}{83.85} / \colorbox{orange!20}{70.17} & \colorbox{orange!20}{88.50} / \colorbox{orange!20}{81.87} \\
\textbf{Hy-MT3-7B} & 92.30 / 89.61 & 93.64 / 90.66 & 90.26 / 88.01 & 88.38 / 84.29 & 90.35 / 85.36 & 81.74 / 79.82 & 82.79 / 68.66 & 86.12 / 80.39 \\
\textbf{Hy-MT3-1.8B} & 85.89 / 85.02 & 88.18 / 86.58 & 81.82 / 81.65 & 73.48 / 72.28 & 77.40 / 74.14 & 70.62 / 70.39 & 71.72 / 60.88 & 75.05 / 71.66 \\
\hdashline
\textbf{Hy-MT2-30B-A3B} & 90.42 / 90.40 & 92.06 / 90.90 & 88.28 / 88.54 & 63.89 / 66.37 & 67.75 / 67.42 & 73.03 / 75.39 & 79.51 / 69.59 & 76.75 / 75.17 \\
\textbf{Hy-MT2-7B} & 86.52 / 88.65 & 89.95 / 89.43 & 84.51 / 86.66 & 37.75 / 44.24 & 38.81 / 44.14 & 66.30 / 69.20 & 78.88 / 68.82 & 67.10 / 67.29 \\
\textbf{Hy-MT2-1.8B} & 80.78 / 82.42 & 82.81 / 83.72 & 75.56 / 78.41 & 24.44 / 27.50 & 26.88 / 30.09 & 53.59 / 55.84 & 70.50 / 62.24 & 56.51 / 56.46 \\
\hline
{\color{orange}Kimi-K3.1-Pro} & 92.95 / 90.79 & 93.82 / 91.27 & 90.60 / 89.00 & 88.98 / 85.74 & 91.68 / 87.31 & 85.86 / \colorbox{orange!20}{81.60} & 76.90 / 68.54 & 86.01 / 81.47 \\
{\color{orange}DeepSeek-V4.1-Flash} & 93.49 / 90.37 & 94.23 / 90.90 & 90.56 / 87.95 & 90.46 / 85.78 & 92.96 / 87.01 & 83.68 / 79.18 & 70.72 / 62.06 & 84.27 / 78.98 \\
{\color{orange}Qwen3.8-Max} & 93.60 / \colorbox{orange!20}{91.51} & 94.56 / \colorbox{orange!20}{92.12} & 91.82 / \colorbox{orange!20}{89.86} & 91.38 / \colorbox{orange!20}{88.27} & \colorbox{orange!20}{92.99} / \colorbox{orange!20}{88.89} & \colorbox{orange!20}{86.88} / 81.28 & 81.94 / 64.36 & 88.02 / 81.08 \\
{\color{orange}Qwen3.8-Flash-Next} & 86.93 / 87.06 & 88.51 / 87.74 & 79.30 / 80.54 & 83.39 / 83.18 & 86.99 / 84.49 & 80.00 / 79.77 & 60.52 / 61.76 & 76.50 / 76.68 \\
{\color{orange}Qwen3.8-27B} & 83.60 / 84.09 & 85.44 / 84.67 & 74.25 / 75.95 & 73.89 / 75.21 & 78.78 / 77.06 & 72.73 / 72.63 & 49.33 / 54.45 & 68.47 / 70.04 \\
{\color{orange}Qwen3.6-35B-A3B} & 87.28 / 87.73 & 89.35 / 88.73 & 82.92 / 84.51 & 76.16 / 77.83 & 82.15 / 80.92 & 73.20 / 74.40 & 50.28 / 56.64 & 71.55 / 73.84 \\
{\color{orange}Gemma4-31B} & 90.69 / 90.02 & 93.23 / 90.93 & 89.33 / 88.75 & 80.47 / 79.51 & 85.82 / 81.64 & 79.98 / 77.72 & 50.73 / 55.38 & 75.87 / 75.66 \\
{\color{orange}Gemma4-26B-A4B} & 89.59 / 89.08 & 92.55 / 90.48 & 87.83 / 87.72 & 73.16 / 73.34 & 81.66 / 78.14 & 76.92 / 75.96 & 40.63 / 43.94 & 70.79 / 70.90 \\
{\color{orange}Gemma4-E4B} & 84.06 / 84.00 & 88.29 / 86.82 & 79.60 / 80.40 & 55.89 / 57.63 & 68.42 / 65.72 & 65.36 / 64.98 & 31.92 / 38.88 & 59.95 / 61.63 \\
{\color{orange}Gemma4-E2B} & 76.39 / 76.95 & 84.88 / 83.56 & 75.30 / 76.20 & 40.61 / 42.45 & 59.12 / 56.38 & 58.13 / 58.28 & 26.78 / 33.18 & 52.68 / 54.39 \\
\hline
{\color{purple}Index-Translate-35B-A3B} & 90.61 / 90.05 & 93.00 / 90.72 & 88.46 / 87.95 & 85.51 / 83.84 & 88.96 / 85.12 & 81.66 / 81.53 & 57.24 / 62.71 & 78.75 / 79.24 \\
{\color{purple}Index-Translate-9B} & 90.13 / 89.68 & 92.49 / 90.36 & 87.30 / 87.13 & 84.47 / 82.84 & 87.95 / 84.62 & 79.86 / 80.64 & 56.80 / 62.23 & 77.66 / 78.52 \\
{\color{purple}Index-Translate-2B} & 76.49 / 77.30 & 78.68 / 77.54 & 63.45 / 63.87 & 64.68 / 65.20 & 68.05 / 66.65 & 64.54 / 66.25 & 44.42 / 50.80 & 60.11 / 62.12 \\
{\color{purple}Tower-Plus-72B} & 79.59 / 78.30 & 80.29 / 78.43 & 67.19 / 66.37 & 52.23 / 51.15 & 55.82 / 53.04 & 65.57 / 64.06 & 27.62 / 28.45 & 53.98 / 53.10 \\
{\color{purple}Qwen-MT-Flash} & 88.70 / 88.98 & 90.06 / 89.15 & 84.72 / 85.71 & 71.03 / 73.89 & 72.43 / 73.56 & 77.77 / 80.05 & 33.21 / 38.37 & 67.00 / 69.56 \\
{\color{purple}Microsoft-Translator} & 85.74 / 85.61 & 89.12 / 87.50 & 83.70 / 83.91 & 76.40 / 76.59 & 79.38 / 78.35 & 66.53 / 66.96 & 55.82 / 50.99 & 71.09 / 69.91 \\
{\color{purple}iFlytek-Translator} & 86.79 / 85.25 & 87.14 / 84.98 & 77.61 / 78.22 & 77.84 / 76.28 & 76.70 / 73.90 & 60.26 / 59.16 & 76.79 / 63.56 & 73.26 / 69.21 \\
{\color{purple}Volcano-Translator} & 83.82 / 82.28 & 86.15 / 84.35 & 80.17 / 79.23 & 51.59 / 52.06 & 52.07 / 52.23 & 51.48 / 51.79 & 30.57 / 36.02 & 53.65 / 54.92 \\
\hline
{\color{blue}GPT-6.1-Sol$^{\mathrm{T}}$} & 94.08 / 91.18 & 95.17 / 91.80 & 92.66 / 89.70 & 93.83 / 88.71 & 93.84 / 88.08 & 89.49 / 81.73 & 92.12 / \colorbox{cyan!20}{71.80} & 92.08 / 82.96 \\
{\color{blue}Claude-Opus-5.5$^{\mathrm{T}}$} & 94.07 / \colorbox{cyan!20}{92.29} & 95.05 / \colorbox{cyan!20}{93.21} & 92.83 / \colorbox{cyan!20}{90.91} & 93.92 / \colorbox{cyan!20}{90.06} & 95.13 / \colorbox{cyan!20}{90.83} & 89.19 / 81.27 & 90.62 / 70.17 & 91.84 / \colorbox{cyan!20}{83.25} \\
{\color{blue}Gemini-4-Argon$^{\mathrm{T}}$} & 93.58 / 91.79 & 94.60 / 92.33 & 91.85 / 90.41 & 93.12 / 89.23 & 94.12 / 89.41 & 87.10 / 79.19 & 86.85 / 69.80 & 89.92 / 82.23 \\
{\color{blue}Kimi-K3.1-Pro$^{\mathrm{T}}$} & 95.34 / 92.08 & 95.82 / 92.47 & 93.35 / 90.50 & 94.32 / 89.04 & 94.80 / 89.41 & 89.46 / 83.04 & 90.88 / 68.10 & 92.13 / 82.77 \\
{\color{blue}DeepSeek-V4.1-Flash$^{\mathrm{T}}$} & \colorbox{cyan!20}{96.19} / 91.70 & \colorbox{cyan!20}{96.75} / 92.14 & \colorbox{cyan!20}{95.08} / 90.04 & \colorbox{cyan!20}{96.14} / 88.89 & \colorbox{cyan!20}{96.61} / 89.37 & \colorbox{cyan!20}{92.57} / 83.46 & \colorbox{cyan!20}{92.18} / 67.87 & \colorbox{cyan!20}{94.09} / 82.68 \\
{\color{blue}GLM-5.3$^{\mathrm{T}}$} & 94.89 / 91.80 & 95.42 / 91.87 & 92.95 / 89.60 & 93.48 / 88.19 & 94.17 / 88.25 & 78.35 / 75.97 & 71.40 / 60.21 & 84.20 / 78.57 \\
{\color{blue}Qwen3.8-Max$^{\mathrm{T}}$} & 94.97 / 91.76 & 96.16 / 92.05 & 94.13 / 90.31 & 94.55 / 89.05 & 95.86 / 89.53 & 91.34 / \colorbox{cyan!20}{83.51} & 84.93 / 66.92 & 91.45 / 82.56 \\
{\color{blue}Qwen3.8-Flash-Next$^{\mathrm{T}}$} & 92.62 / 90.24 & 94.27 / 90.76 & 90.79 / 88.07 & 91.52 / 87.39 & 92.84 / 87.27 & 86.67 / 80.15 & 80.15 / 60.60 & 87.52 / 79.11 \\
{\color{blue}Qwen3.8-27B$^{\mathrm{T}}$} & 90.31 / 88.32 & 91.70 / 88.30 & 86.22 / 84.34 & 84.80 / 82.04 & 86.20 / 81.37 & 78.84 / 73.48 & 64.10 / 53.71 & 78.81 / 73.43 \\
{\color{blue}Qwen3.6-35B-A3B$^{\mathrm{T}}$} & 91.83 / 90.24 & 93.00 / 90.50 & 88.90 / 87.96 & 87.54 / 84.78 & 88.68 / 84.47 & 76.32 / 74.79 & 66.03 / 59.12 & 79.94 / 76.70 \\
{\color{blue}Gemma4-31B$^{\mathrm{T}}$} & 93.45 / 91.20 & 94.94 / 91.73 & 92.39 / 89.90 & 88.35 / 85.52 & 90.48 / 85.77 & 84.10 / 80.17 & 73.12 / 55.26 & 84.81 / 77.79 \\
{\color{blue}Gemma4-26B-A4B$^{\mathrm{T}}$} & 93.01 / 90.86 & 94.67 / 91.54 & 91.91 / 89.34 & 85.22 / 82.94 & 87.74 / 83.42 & 82.09 / 80.15 & 60.73 / 47.61 & 80.36 / 75.13 \\
{\color{blue}Gemma4-E4B$^{\mathrm{T}}$} & 84.10 / 84.57 & 89.05 / 87.51 & 83.00 / 83.43 & 58.22 / 59.79 & 69.79 / 67.36 & 66.21 / 66.08 & 32.80 / 38.86 & 61.61 / 63.06 \\
{\color{blue}Gemma4-E2B$^{\mathrm{T}}$} & 85.05 / 84.57 & 87.77 / 86.39 & 81.89 / 82.14 & 59.83 / 59.87 & 64.10 / 61.46 & 60.74 / 61.38 & 32.15 / 41.25 & 59.32 / 61.46 \\
\bottomrule
\end{tabular}
}
\vspace{1mm}

\begin{minipage}{0.98\textwidth}
\tiny
\textit{Notes.} Each cell reports TransRubric / GEMBA scores on a 0--100 scale; higher is better. Both are reference-based LLM-as-a-judge metrics. TransRubric is the arithmetic mean of two pointwise judges, DeepSeek-V4.1-Flash and Qwen3.5-397B-A17B, with scores averaged over the fixed expected number of samples. GEMBA uses GPT-OSS-120B with medium reasoning and is macro-averaged over language pairs. For both metrics, empty translations receive a score of 0.
For the 33-language FLORES-200 set, ZH$\Leftrightarrow$XX covers all directions involving Chinese, including Chinese--English; EN$\Leftrightarrow$XX covers directions involving English but not Chinese; and XX$\Leftrightarrow$XX covers directions involving neither Chinese nor English. \textbf{Mand.}$\Leftrightarrow$\textbf{Min.} denotes Mandarin$\Leftrightarrow$Minority translation. \textbf{Avg.} is the unweighted mean of four benchmark-level scores: FLORES-200, FLORES-200 low-resource, WMT26 and Mand.$\Leftrightarrow$Min.
$^{\mathrm{T}}$ denotes thinking mode; unmarked models use non-thinking mode. Our models are shown in bold. \colorbox{orange!20}{Orange} and \colorbox{cyan!20}{blue} backgrounds indicate the best non-thinking and thinking results, respectively, in each metric column.
\end{minipage}
\end{table*}

%% file: tables/wild_hard_if.tex
\begin{table*}[!htbp]
\centering
\caption{Performance comparison on in-the-wild, hard-case and instruction-following translation benchmarks.}
\label{tab:wild_hard_if}
\scriptsize
\renewcommand{\arraystretch}{1.18}
\setlength{\tabcolsep}{4pt}
\resizebox{0.85\textwidth}{!}{
\begin{tabular}{lccccc}
\toprule
\multirow{2}{*}{\textbf{Models}}
& \multirow{2}{*}{\textbf{WildMTBench-202611}}
& \multirow{2}{*}{\textbf{HardMTBench}}
& \multicolumn{3}{c}{\textbf{IFMTBench}} \\
\cmidrule(lr){4-6}
& & & Simple & Complex & Total \\
\midrule
\textbf{Hy-MT3-30B-A3B} & \colorbox{orange!20}{91.30} & 88.93 / 89.80 & \colorbox{orange!20}{91.02} & 79.74 & \colorbox{orange!20}{86.66} \\
\textbf{Hy-MT3-7B} & 89.09 & 83.91 / 86.50 & 88.36 & 76.19 & 83.66 \\
\textbf{Hy-MT3-1.8B} & 84.45 & 76.44 / 81.77 & 76.40 & 55.56 & 68.34 \\
\hdashline
\textbf{Hy-MT2-30B-A3B} & 83.62 & 81.38 / 88.84 & 89.14 & 74.14 & 83.35 \\
\textbf{Hy-MT2-7B} & 80.99 & 73.16 / 84.12 & 87.87 & 71.08 & 81.38 \\
\textbf{Hy-MT2-1.8B} & 76.91 & 68.07 / 78.15 & 74.76 & 52.68 & 66.23 \\
\hline
{\color{orange}Kimi-K3.1-Pro} & 90.23 & 92.04 / 91.47 & 85.53 & 82.21 & 84.25 \\
{\color{orange}DeepSeek-V4.1-Flash} & 89.82 & 90.48 / 90.34 & 81.75 & 79.06 & 80.71 \\
{\color{orange}Qwen3.8-Max} & 90.94 & \colorbox{orange!20}{92.30} / \colorbox{orange!20}{91.71} & 88.79 & \colorbox{orange!20}{82.84} & 86.49 \\
{\color{orange}Qwen3.8-Flash-Next} & 86.28 & 87.98 / 90.64 & 83.02 & 78.89 & 81.42 \\
{\color{orange}Qwen3.8-27B} & 85.46 & 84.54 / 88.13 & 81.55 & 76.29 & 79.52 \\
{\color{orange}Qwen3.6-35B-A3B} & 84.53 & 84.26 / 88.62 & 77.73 & 70.02 & 74.75 \\
{\color{orange}Gemma4-31B} & 84.77 & 84.74 / 88.31 & 84.25 & 72.77 & 79.81 \\
{\color{orange}Gemma4-26B-A4B} & 85.41 & 83.01 / 87.17 & 82.89 & 73.59 & 79.29 \\
{\color{orange}Gemma4-E4B} & 82.11 & 74.23 / 81.17 & 74.19 & 67.05 & 71.43 \\
{\color{orange}Gemma4-E2B} & 76.90 & 69.27 / 76.94 & 62.86 & 48.94 & 57.48 \\
\hline
{\color{purple}Index-Translate-35B-A3B} & 86.47 & 85.57 / 88.94 & 80.26 & 76.88 & 78.96 \\
{\color{purple}Index-Translate-9B} & 85.94 & 83.45 / 88.26 & 78.49 & 72.76 & 76.27 \\
{\color{purple}Index-Translate-2B} & 77.56 & 73.06 / 80.72 & 70.66 & 57.67 & 65.64 \\
{\color{purple}Tower-Plus-72B} & 84.55 & 82.12 / 87.34 & 79.01 & 73.07 & 76.72 \\
{\color{purple}Qwen-MT-Flash} & -- & 84.22 / 88.74 & -- & -- & -- \\
{\color{purple}Microsoft-Translator} & -- & 71.13 / 77.21 & -- & -- & -- \\
{\color{purple}iFlytek-Translator} & -- & 73.99 / 78.81 & -- & -- & -- \\
{\color{purple}Volcano-Translator} & -- & 69.43 / 73.66 & -- & -- & -- \\
\hline
{\color{blue}GPT-6.1-Sol$^{\mathrm{T}}$} & 92.50 & 93.26 / 91.98 & 90.68 & \colorbox{cyan!20}{84.75} & 88.38 \\
{\color{blue}Claude-Opus-5.5$^{\mathrm{T}}$} & 92.09 & 92.90 / 91.85 & 87.44 & 75.51 & 82.82 \\
{\color{blue}Gemini-4-Argon$^{\mathrm{T}}$} & 90.30 & 91.65 / 91.32 & \colorbox{cyan!20}{93.41} & 83.73 & \colorbox{cyan!20}{89.68} \\
{\color{blue}Kimi-K3.1-Pro$^{\mathrm{T}}$} & 93.51 & \colorbox{cyan!20}{94.69} / \colorbox{cyan!20}{92.30} & 88.89 & \colorbox{cyan!20}{84.75} & 87.29 \\
{\color{blue}DeepSeek-V4.1-Flash$^{\mathrm{T}}$} & \colorbox{cyan!20}{94.21} & 94.38 / 91.53 & 86.57 & 83.46 & 85.37 \\
{\color{blue}GLM-5.3$^{\mathrm{T}}$} & 91.39 & 92.25 / 90.75 & 86.85 & 81.29 & 84.70 \\
{\color{blue}Qwen3.8-Max$^{\mathrm{T}}$} & 91.96 & 93.49 / 91.72 & 87.87 & 83.87 & 86.32 \\
{\color{blue}Qwen3.8-Flash-Next$^{\mathrm{T}}$} & 90.40 & 91.11 / 91.13 & 88.09 & 84.06 & 86.53 \\
{\color{blue}Qwen3.8-27B$^{\mathrm{T}}$} & 90.55 & 89.24 / 89.45 & 88.00 & 83.49 & 86.26 \\
{\color{blue}Qwen3.6-35B-A3B$^{\mathrm{T}}$} & 88.73 & 88.00 / 90.55 & 84.02 & 82.42 & 83.40 \\
{\color{blue}Gemma4-31B$^{\mathrm{T}}$} & 88.91 & 87.28 / 89.54 & 86.11 & 74.21 & 81.51 \\
{\color{blue}Gemma4-26B-A4B$^{\mathrm{T}}$} & 88.70 & 86.23 / 88.77 & 84.48 & 74.61 & 80.67 \\
{\color{blue}Gemma4-E4B$^{\mathrm{T}}$} & 82.01 & 75.06 / 82.28 & 76.84 & 69.65 & 74.06 \\
{\color{blue}Gemma4-E2B$^{\mathrm{T}}$} & 80.71 & 71.54 / 78.90 & 73.87 & 63.97 & 70.04 \\
\bottomrule
\end{tabular}
}
\vspace{1mm}

\begin{minipage}{0.98\textwidth}
\tiny
\textit{Notes.} All scores are on a 0--100 scale; higher is better. TransRubric follows the same two-judge protocol as in Table~\ref{tab:general_translation}, with empty or missing outputs scored as 0. \textbf{WildMTBench-202611} contains 5{,}000 real-world samples and reports reference-free TransRubric scores using the fixed expected number of samples as the denominator. \textbf{HardMTBench} contains bidirectional Chinese--English cases and reports TransRubric / GEMBA scores, with GEMBA macro-averaged over the two directions. \textbf{IFMTBench} is evaluated using its official rules and GPT-OSS-120B judge and reports the mean over valid samples. \textit{Simple} and \textit{Complex} denote the single- and multi-constraint subsets, respectively, and \textit{Total} aggregates all 7{,}344 test cases across both subsets.
$^{\mathrm{T}}$ denotes thinking mode; unmarked models use non-thinking mode. A dash indicates that the model was not evaluated on the corresponding benchmark. Our models are shown in bold. \colorbox{orange!20}{Orange} and \colorbox{cyan!20}{blue} backgrounds indicate the best non-thinking and thinking results, respectively, in each score column.
\end{minipage}
\end{table*}

%% file: tables/wild_domain.tex
\begin{table*}[!htbp]
\centering
\caption{Domain-wise performance on WildMTBench-202611.}
\label{tab:wild_domain}
\scriptsize
\renewcommand{\arraystretch}{1.18}
\setlength{\tabcolsep}{4pt}
\resizebox{\textwidth}{!}{
\begin{tabular}{lccccccccccc}
\toprule
\textbf{Models} & \textbf{Politics} & \textbf{Finance} & \textbf{Tech.} & \textbf{Medical} & \textbf{Legal} & \textbf{Education} & \textbf{Culture \& Ent.} & \textbf{Sports} & \textbf{Life \& Social} & \textbf{Others} & \textbf{Avg.} \\
\midrule
\textbf{Hy-MT3-30B-A3B} & \colorbox{orange!20}{91.80} & \colorbox{orange!20}{91.41} & \colorbox{orange!20}{92.63} & \colorbox{orange!20}{92.32} & 92.57 & 89.73 & 91.55 & 90.17 & 89.90 & \colorbox{orange!20}{90.93} & \colorbox{orange!20}{91.30} \\
\textbf{Hy-MT3-7B} & 89.87 & 89.30 & 90.21 & 90.02 & 91.69 & 87.81 & 88.74 & 86.73 & 87.91 & 88.61 & 89.09 \\
\textbf{Hy-MT3-1.8B} & 84.45 & 84.26 & 86.66 & 85.38 & 87.43 & 83.89 & 83.72 & 80.16 & 82.89 & 85.64 & 84.45 \\
\hdashline
\textbf{Hy-MT2-30B-A3B} & 84.17 & 84.17 & 83.72 & 81.33 & 86.15 & 82.14 & 84.13 & 83.86 & 83.01 & 83.49 & 83.62 \\
\textbf{Hy-MT2-7B} & 81.19 & 82.14 & 80.95 & 79.11 & 83.39 & 80.00 & 81.95 & 79.70 & 81.00 & 80.52 & 80.99 \\
\textbf{Hy-MT2-1.8B} & 76.73 & 78.11 & 77.97 & 75.14 & 80.99 & 76.47 & 76.94 & 73.41 & 73.93 & 79.38 & 76.91 \\
\hline
{\color{orange}Kimi-K3.1-Pro} & 90.78 & 89.51 & 91.45 & 91.12 & 91.86 & 89.66 & 90.48 & 90.84 & 88.53 & 88.04 & 90.23 \\
{\color{orange}DeepSeek-V4.1-Flash} & 89.85 & 90.32 & 90.96 & 90.89 & 91.82 & 89.36 & 89.61 & 90.21 & 88.30 & 86.91 & 89.82 \\
{\color{orange}Qwen3.8-Max} & 90.90 & 91.33 & 92.33 & 91.59 & \colorbox{orange!20}{92.74} & \colorbox{orange!20}{90.05} & \colorbox{orange!20}{91.61} & \colorbox{orange!20}{91.14} & \colorbox{orange!20}{90.03} & 87.70 & 90.94 \\
{\color{orange}Qwen3.8-Flash-Next} & 87.38 & 86.22 & 88.23 & 86.99 & 88.42 & 85.72 & 86.32 & 86.09 & 83.72 & 83.68 & 86.28 \\
{\color{orange}Qwen3.8-27B} & 86.11 & 86.21 & 87.77 & 85.83 & 88.18 & 84.96 & 85.21 & 84.76 & 82.68 & 82.94 & 85.46 \\
{\color{orange}Qwen3.6-35B-A3B} & 85.24 & 85.96 & 86.36 & 85.49 & 87.14 & 83.84 & 83.87 & 83.88 & 81.54 & 81.97 & 84.53 \\
{\color{orange}Gemma4-31B} & 86.63 & 80.29 & 86.00 & 85.21 & 87.88 & 83.61 & 87.20 & 87.40 & 79.96 & 83.48 & 84.77 \\
{\color{orange}Gemma4-26B-A4B} & 86.21 & 85.75 & 86.80 & 85.98 & 88.16 & 84.22 & 86.26 & 84.88 & 82.56 & 83.24 & 85.41 \\
{\color{orange}Gemma4-E4B} & 81.37 & 82.74 & 84.17 & 81.12 & 84.77 & 82.29 & 82.19 & 78.75 & 82.29 & 81.36 & 82.11 \\
{\color{orange}Gemma4-E2B} & 74.80 & 76.62 & 80.76 & 76.22 & 81.29 & 78.59 & 77.43 & 68.42 & 77.16 & 77.73 & 76.90 \\
\hline
{\color{purple}Index-Translate-35B-A3B} & 84.57 & 87.14 & 87.97 & 86.22 & 87.65 & 86.12 & 87.06 & 85.51 & 86.44 & 85.97 & 86.47 \\
{\color{purple}Index-Translate-9B} & 84.48 & 86.91 & 87.14 & 85.49 & 88.05 & 85.36 & 87.06 & 84.34 & 85.88 & 84.69 & 85.94 \\
{\color{purple}Index-Translate-2B} & 76.05 & 79.68 & 79.84 & 77.84 & 81.55 & 77.73 & 76.39 & 73.15 & 75.30 & 78.10 & 77.56 \\
{\color{purple}Tower-Plus-72B} & 84.36 & 85.70 & 85.91 & 84.72 & 86.86 & 84.52 & 84.22 & 83.84 & 81.99 & 83.38 & 84.55 \\
\hline
{\color{blue}GPT-6.1-Sol$^{\mathrm{T}}$} & 91.71 & 91.85 & 92.70 & 93.28 & 93.71 & 91.62 & 93.09 & 93.67 & 91.26 & 92.10 & 92.50 \\
{\color{blue}Claude-Opus-5.5$^{\mathrm{T}}$} & 92.54 & 91.91 & 92.18 & 92.02 & 93.40 & 91.51 & 91.91 & 93.01 & 91.87 & 90.53 & 92.09 \\
{\color{blue}Gemini-4-Argon$^{\mathrm{T}}$} & 91.08 & 89.92 & 90.91 & 92.16 & 91.80 & 88.87 & 89.71 & 91.03 & 89.89 & 87.60 & 90.30 \\
{\color{blue}Kimi-K3.1-Pro$^{\mathrm{T}}$} & 93.56 & 92.78 & 94.46 & 94.16 & 94.66 & 92.88 & 93.49 & 94.12 & 93.20 & 91.75 & 93.51 \\
{\color{blue}DeepSeek-V4.1-Flash$^{\mathrm{T}}$} & \colorbox{cyan!20}{94.31} & \colorbox{cyan!20}{93.41} & \colorbox{cyan!20}{94.86} & \colorbox{cyan!20}{94.58} & \colorbox{cyan!20}{95.59} & \colorbox{cyan!20}{93.08} & \colorbox{cyan!20}{94.81} & \colorbox{cyan!20}{94.89} & \colorbox{cyan!20}{93.39} & \colorbox{cyan!20}{93.15} & \colorbox{cyan!20}{94.21} \\
{\color{blue}GLM-5.3$^{\mathrm{T}}$} & 90.84 & 90.92 & 92.28 & 91.95 & 93.28 & 90.96 & 91.98 & 92.03 & 90.24 & 89.47 & 91.39 \\
{\color{blue}Qwen3.8-Max$^{\mathrm{T}}$} & 91.58 & 92.04 & 93.29 & 92.73 & 93.32 & 91.62 & 92.34 & 92.65 & 91.31 & 88.75 & 91.96 \\
{\color{blue}Qwen3.8-Flash-Next$^{\mathrm{T}}$} & 90.06 & 90.90 & 92.30 & 91.77 & 91.57 & 90.02 & 90.39 & 89.41 & 88.70 & 88.94 & 90.40 \\
{\color{blue}Qwen3.8-27B$^{\mathrm{T}}$} & 91.00 & 90.36 & 92.81 & 91.22 & 92.14 & 89.61 & 90.18 & 90.13 & 88.76 & 89.31 & 90.55 \\
{\color{blue}Qwen3.6-35B-A3B$^{\mathrm{T}}$} & 88.67 & 87.99 & 90.12 & 90.53 & 90.74 & 87.60 & 88.98 & 88.93 & 86.32 & 87.46 & 88.73 \\
{\color{blue}Gemma4-31B$^{\mathrm{T}}$} & 89.93 & 87.06 & 89.63 & 89.12 & 91.06 & 87.28 & 90.71 & 89.74 & 86.58 & 88.00 & 88.91 \\
{\color{blue}Gemma4-26B-A4B$^{\mathrm{T}}$} & 89.01 & 87.82 & 89.41 & 89.45 & 89.91 & 87.03 & 89.56 & 89.30 & 87.30 & 88.16 & 88.70 \\
{\color{blue}Gemma4-E4B$^{\mathrm{T}}$} & 81.93 & 82.52 & 83.70 & 82.06 & 84.79 & 82.22 & 82.06 & 78.90 & 81.31 & 80.61 & 82.01 \\
{\color{blue}Gemma4-E2B$^{\mathrm{T}}$} & 78.22 & 82.08 & 85.02 & 80.84 & 84.25 & 81.64 & 80.18 & 72.05 & 81.59 & 81.26 & 80.71 \\
\bottomrule
\end{tabular}
}
\vspace{1mm}

\begin{minipage}{0.98\textwidth}
\tiny
\textit{Notes.} All scores are reference-free TransRubric scores on a 0--100 scale; higher is better. \textbf{WildMTBench-202611} contains 5{,}000 samples evenly divided into 10 domains of 500 samples each. Each domain score is the mean of the DeepSeek-V4.1-Flash and Qwen3.5-397B-A17B judge scores, using 500 as the fixed denominator; empty or missing translations are assigned a score of 0. \textbf{Avg.} is the score over all 5{,}000 samples and, because the domains are equally sized, is also the unweighted mean of the 10 domain scores. Judge-side failures are also scored as 0 (one case each for Hy-MT3-30B-A3B and Hy-MT3-7B, and two cases for Claude-Opus-5.5$^{\mathrm{T}}$).
$^{\mathrm{T}}$ denotes thinking mode; unmarked models use non-thinking mode. Commercial translation APIs were not evaluated and are omitted. \colorbox{orange!20}{Orange} and \colorbox{cyan!20}{blue} backgrounds indicate the best non-thinking and thinking results, respectively, in each column.
\end{minipage}
\end{table*}

%% file: tables/wild_breakdown.tex
\begin{table}[!htbp]
\centering
\caption{Fine-grained comparison between Hy-MT2-30B-A3B and Hy-MT3-30B-A3B on WildMTBench-202611.}
\label{tab:wild_breakdown}
\scriptsize
\renewcommand{\arraystretch}{1.18}
\setlength{\tabcolsep}{5pt}
\begin{tabular}{lccc}
\toprule
\multicolumn{4}{l}{\textit{(a) Score distribution}} \\
\midrule
\textbf{Metric} & \textbf{Hy-MT2-30B-A3B} & \textbf{Hy-MT3-30B-A3B} & \textbf{Change} \\
\midrule
Average judge score $\uparrow$ & 83.62 & \textbf{91.30} & +7.68 \\
Perfect-score rate (score $=100$) $\uparrow$ & 25.5\% & \textbf{40.6\%} & +15.1 pt \\
Samples with score $\geq 90$ $\uparrow$ & 46.1\% & \textbf{70.9\%} & +24.8 pt \\
Samples with score $\geq 80$ $\uparrow$ & 68.2\% & \textbf{87.2\%} & +19.0 pt \\
Poor-translation rate (score $<60$) $\downarrow$ & 7.4\% & \textbf{2.2\%} & $-$5.2 pt ($-$70\%) \\
Average issues per sample $\downarrow$ & 2.14 & \textbf{1.13} & $-$1.01 ($-$47\%) \\
\midrule
\multicolumn{4}{l}{\textit{(b) Error types (share of samples, $\downarrow$)}} \\
\midrule
\textbf{Error type} & \textbf{Hy-MT2-30B-A3B} & \textbf{Hy-MT3-30B-A3B} & \textbf{Rel. change} \\
\midrule
Completeness (omission / addition) & 21.2\% & \textbf{5.5\%} & $-$74\% \\
Accuracy (mistranslation) & 56.0\% & \textbf{31.8\%} & $-$43\% \\
Terminology & 28.3\% & \textbf{17.9\%} & $-$37\% \\
Instruction following & 16.5\% & \textbf{9.8\%} & $-$40\% \\
\bottomrule
\end{tabular}
\vspace{1mm}
\begin{minipage}{0.98\linewidth}
\tiny
\textit{Notes.} All statistics are computed over the 5{,}000 WildMTBench-202611 samples using TransRubric scores and issue annotations.
$\uparrow$ and $\downarrow$ indicate that higher and lower values are better, respectively.
``pt'' denotes an absolute change in percentage points, while values in parentheses denote relative changes from Hy-MT2-30B-A3B.
In (b), each value is the share of samples containing at least one issue of the corresponding type; issue types are not mutually exclusive.
\end{minipage}
\end{table}

%% file: tables/case_study.tex
\begin{figure}[!htbp]
\centering
\caption{Case study comparing Hy-MT2 and Hy-MT3.}
\label{fig:case_study}
\definecolor{mtorange}{RGB}{255,128,0}
\definecolor{mtred}{RGB}{255,0,0}
\definecolor{mtblue}{RGB}{0,0,255}
\definecolor{mtpurple}{RGB}{128,0,128}
\newcommand{\cons}[1]{\textcolor{mtorange}{#1}}
\newcommand{\instr}[1]{\textcolor{mtpurple}{#1}}
\newcommand{\bad}[1]{\textcolor{mtred}{#1}}
\newcommand{\good}[1]{\textcolor{mtblue}{#1}}
\newcommand{\xtag}[1]{\textless #1\textgreater}
\newcommand{\grouptype}[1]{\multirow{2}{=}{\raggedright\scriptsize #1}}
\newcommand{\cellsep}{\par\vspace{3pt}}
\newcolumntype{T}{>{\raggedright\arraybackslash\scriptsize}p{0.08\linewidth}}
\newcolumntype{P}{>{\raggedright\arraybackslash\scriptsize}p{0.285\linewidth}}
\newcolumntype{M}{>{\raggedright\arraybackslash\scriptsize}p{0.28\linewidth}}
\setlength{\tabcolsep}{4pt}
\renewcommand{\arraystretch}{1.45}
\setlength{\aboverulesep}{7pt}
\setlength{\belowrulesep}{7pt}
\sloppy

\begin{tabular}{T P M M}
\toprule
\textbf{Type} & \textbf{Prompt} & \textbf{Hy-MT2 Response} & \textbf{Hy-MT3 Response} \\
\midrule

\grouptype{Mistrans\-lation}
&
将以下文本翻译为英语，注意只需要输出翻译后的结果，不要额外解释：
\cellsep
\cons{夏雨雨}感动无数国人，14亿人为之\cons{动容} ｜ 体坛记忆
&
The \bad{summer rains} touched the hearts of countless Chinese people;
1.4 billion individuals were \bad{moved to tears} | Sports Memories
&
\good{Xia Yuyu} Moved Countless Chinese People---1.4 Billion \good{Were Touched} | Sports Memories
\\
\cmidrule(lr){2-4}

&
将以下文本从英语翻译为简体中文，注意只需要输出翻译后的结果，不要额外解释：%
Creatine MagnaPower\textregistered{} unites the \cons{powerful pairing} of creatine and magnesium
in a clinically proven formulation \cons{designed to support} strength, power and high-intensity performance.
&
Creatine MagnaPower\textregistered{}将\bad{具有卓越功效的} creatine与镁元素\bad{完美}结合，
这种经过临床验证的配方\bad{有助于提升}力量、爆发力以及高强度运动表现。
&
Creatine MagnaPower\textregistered{} 将肌酸与镁这一\good{强效组合}融合于经临床验证的配方之中，
该配方\good{旨在助力提升}力量、爆发力及高强度表现。
\\
\midrule

\grouptype{Omission}
&
将以下文本翻译为简体中文，注意只需要输出翻译后的结果，不要额外解释：
\cellsep
\xtag{x0}Men's Hockey Debuts Propel No. 13 Men's Hockey to a \cons{Chippy} \cons{Exhibition}
Victory Over University of Waterloo\xtag{/x0}
&
\xtag{x0}男子冰球新人的出色表现助力排名第13的男子冰球队\bad{以险胜}战胜滑铁卢大学\xtag{/x0}
&
\xtag{x0}男子冰球首秀助力排名第13的男子冰球队在一场\good{对抗激烈的表演赛}中战胜滑铁卢大学\xtag{/x0}
\\
\cmidrule(lr){2-4}

&
将以下文本翻译为英语，注意只需要输出翻译后的结果，不要额外解释：
\cellsep
野球场\cons{硬核}投篮，肘姐出手就进，这准度谁能顶得住
&
In a streetball game, Elbow Sister made every shot she took;
no one could match such accuracy.
&
\good{Hardcore shots} on the street court---Sister Elbow drains it the moment she releases---who
can handle this kind of accuracy?
\\
\midrule

\grouptype{Instruction following}
&
将以下文本翻译为中文，注意只需要输出翻译后的结果，不要额外解释：
\cellsep
\cons{Wang R, Liang L, Matsumoto M, Iwata K, Umemura A, He F.}
Reactive oxygen species and NRF2 signaling, friends or foes in cancer?
Biomolecules 13: 353, 2023.
&
\bad{王R、梁L、松本M、岩田K、梅村A、何F。}活性氧与NRF2信号通路：在癌症中是“帮手”还是“敌人”？
《生物分子》13: 353，2023年。
&
\good{Wang R, Liang L, Matsumoto M, Iwata K, Umemura A, He F.}
活性氧与NRF2信号通路：在癌症中是盟友还是对手？Biomolecules 13: 353, 2023.
\\
\cmidrule(lr){2-4}

&
\instr{用最符合口语表达的标准进行翻译}
\cellsep
参考上面的信息，把下面的文本翻译成英语，注意不需要翻译上文，也不要额外解释：
\cellsep
我们要特别重视挖掘中华五千年文明中的精华，把弘扬优秀文化同马克思主义立场观点方法
&
\bad{We must pay special attention to drawing out the essence of China's 5,000-year civilization,
and integrate the promotion of excellent culture with the stance, viewpoints, and methods of Marxism.}
&
\good{We really need to pay special attention to drawing out the best of China's
five-thousand-year-old civilization, and to linking the promotion of fine culture with the
positions, viewpoints, and methods of Marxism}
\\
\bottomrule
\end{tabular}
\end{figure}

%% file: tables/game.tex
\begin{table}[!htbp]
\centering
\caption{Human evaluation on game translation.}
\label{tab:game}
\scriptsize
\renewcommand{\arraystretch}{1.18}
\setlength{\tabcolsep}{4pt}
\resizebox{\linewidth}{!}{
\begin{tabular}{lccccc}
\toprule
\textbf{Models} & \textbf{Off-target}$\downarrow$ & \textbf{Placeholder Error}$\downarrow$ & \textbf{Rich-text Error}$\downarrow$ & \textbf{Terminology Error}$\downarrow$ & \textbf{Total Score}$\uparrow$ \\
\midrule
\textbf{Hy-MT3-30B-A3B} & \colorbox{orange!20}{0.24\%} & \colorbox{orange!20}{0.08\%} & \colorbox{orange!20}{0.08\%} & \colorbox{orange!20}{0.08\%} & \colorbox{orange!20}{89.6} \\
\hline
{\color{blue}Gemini 3.1 Pro$^{\mathrm{T}}$} & 0.38\% & \colorbox{orange!20}{0.08\%} & 0.19\% & 0.54\% & 88.7 \\
{\color{blue}GPT-5.5$^{\mathrm{T}}$} & 0.58\% & 0.15\% & 0.46\% & 0.27\% & 88.6 \\
{\color{blue}DeepSeek-V4-Pro$^{\mathrm{T}}$} & 0.65\% & 0.12\% & 0.46\% & 0.54\% & 88.3 \\
{\color{blue}Qwen3.6-35B-A3B$^{\mathrm{T}}$} & 2.81\% & 0.42\% & 1.38\% & 1.15\% & 83.4 \\
{\color{blue}Gemma4-26B-A4B$^{\mathrm{T}}$} & 2.11\% & 0.35\% & 1.38\% & 0.81\% & 83.2 \\
\bottomrule
\end{tabular}
}
\vspace{1mm}

\begin{minipage}{0.98\linewidth}
\tiny
\textit{Notes.} The evaluation set and the human evaluation protocol are jointly built with the Honor of Kings team. Off-target, placeholder, rich-text and terminology errors are reported as the percentage of samples containing the corresponding error; lower is better ($\downarrow$). The total score is the overall human rating; higher is better ($\uparrow$).
$^{\mathrm{T}}$ denotes thinking mode. Our model is shown in bold and thinking models in {\color{blue}blue}.
The best result in each column is highlighted with an \colorbox{orange!20}{orange background}.
\end{minipage}
\end{table}

%% file: tables/vista.tex
\begin{table}[!htbp]
\centering
\caption{Translation quality on VISTA-Bench (text track, PaddleOCR input).}
\label{tab:vista_bench}
\begin{minipage}[c]{0.40\linewidth}
\centering
\scriptsize
\renewcommand{\arraystretch}{1.18}
\setlength{\tabcolsep}{6pt}
\begin{tabular}{lc}
\toprule
\textbf{Models} & \textbf{Translation quality} \\
\midrule
\textbf{Hy-MT3-30B-A3B} & \colorbox{orange!20}{45.73} \\
\hline
{\color{orange}Gemma4-26B-A4B} & 45.46 \\
{\color{orange}Qwen3.6-35B-A3B} & 41.82 \\
\hline
{\color{purple}Index-Translate-35B-A3B} & 43.63 \\
{\color{purple}Index-Translate-9B} & 43.05 \\
{\color{purple}Index-Translate-2B} & 33.35 \\
\hline
{\color{blue}GPT-6.1-Sol$^{\mathrm{T}}$} & \colorbox{cyan!20}{51.97} \\
{\color{blue}GLM-5.3$^{\mathrm{T}}$} & 42.84 \\
\bottomrule
\end{tabular}
\end{minipage}\hfill
\begin{minipage}[c]{0.57\linewidth}
\tiny
\textit{Notes.} All scores are on a 0--100 scale.
Test data and the judging protocol are taken from VISTA-Bench~\citep{lv2026vistabenchbenchmarkingmultilingualimage}, a multilingual image translation benchmark covering 23 source languages and 10 everyday-use domains with image-specific rubrics.
We evaluate in a cascaded setting: text in each test image is first recognized by PaddleOCR, and the recognized text is then given to each model for translation, so no model has access to the image.
Following the official protocol, a text-only LLM judge checks each translation against the image-specific rubric without seeing the image or the reference translation; each atomic criterion is scored as fully met or unmet, criteria are weighted by importance (core / supporting / minor $=3/2/1$), and we report the translation-quality (T) dimension averaged over all 506 translation directions.
Because OCR errors (misrecognized, missing or wrongly segmented text) propagate to the translation input and cannot be recovered without the image, the absolute scores of all models are considerably lower than on text-only benchmarks; the scores are intended for relative comparison under identical OCR input.
$^{\mathrm{T}}$ denotes thinking mode; models without the marker are run in non-thinking mode.
Our model is shown in bold at the top, followed by
{\color{orange}general models in non-thinking mode},
{\color{purple}translation-specialized models}, and
{\color{blue}models in thinking mode}.
The best result among non-thinking-mode systems is highlighted with an \colorbox{orange!20}{orange background},
and the best result among {\color{blue}thinking-mode} models with a \colorbox{cyan!20}{blue background}.
\end{minipage}
\end{table}

%% file: tables/quantization.tex
\begin{table*}[!htbp]
\centering
\caption{Performance comparison of Hy-MT3 models under different quantization settings.}
\label{tab:quantization_results}
\scriptsize
\renewcommand{\arraystretch}{1.18}
\setlength{\tabcolsep}{6pt}
\resizebox{\textwidth}{!}{
\begin{tabular}{lccccccc}
\toprule
\multirow{2}{*}{\textbf{Models}}
& \multicolumn{2}{c}{\textbf{FLORES-200}}
& \multirow{2}{*}{\textbf{WMT26}}
& \multirow{2}{*}{\textbf{Mand}}
& \multirow{2}{*}{\textbf{WMTB}}
& \multirow{2}{*}{\textbf{HMTB}}
& \multirow{2}{*}{\textbf{IFMTB}} \\
\cmidrule(lr){2-3}
& \textbf{All-33}
& \textbf{Low-Res.}
& & & & & \\
\midrule
Hy-MT3-1.8B-BF16 & 82.44 & 75.44 & 70.62 & 71.72 & 84.45 & 76.44 & 68.34 \\
Hy-MT3-1.8B-FP8 & 81.41 & 73.61 & 69.08 & 70.16 & 83.88 & 76.09 & 65.09 \\
Hy-MT3-1.8B-Q8 & 82.11 & 74.96 & 70.08 & 71.68 & 84.36 & 76.38 & 66.74 \\
Hy-MT3-1.8B-Q4\_K\_M & 81.16 & 72.89 & 69.00 & 69.20 & 83.89 & 76.01 & 66.01 \\
\hdashline
Hy-MT3-7B-BF16 & 90.58 & 89.36 & 81.74 & 82.79 & 89.09 & 83.91 & 83.66 \\
Hy-MT3-7B-FP8 & 90.48 & 89.31 & 81.63 & 82.36 & 88.86 & 83.73 & 83.06 \\
Hy-MT3-7B-Q8 & 90.50 & 89.26 & 81.67 & 82.56 & 88.94 & 83.87 & 83.21 \\
Hy-MT3-7B-Q4\_K\_M & 90.45 & 89.21 & 81.53 & 82.45 & 88.87 & 83.78 & 83.34 \\
Hy-MT3-7B-IQ2 & 89.59 & 87.57 & 79.85 & 80.84 & 88.34 & 82.81 & 81.85 \\
\hdashline
Hy-MT3-30B-A3B-BF16 & 92.40 & 92.13 & 85.63 & 83.85 & 91.30 & 88.93 & 86.66 \\
Hy-MT3-30B-A3B-FP8 & 92.29 & 92.06 & 85.40 & 83.85 & 91.09 & 88.78 & 86.84 \\
Hy-MT3-30B-A3B-Q8 & 92.37 & 92.15 & 85.65 & 83.98 & 91.17 & 88.93 & 86.69 \\
Hy-MT3-30B-A3B-Q4\_K\_M & 92.33 & 92.09 & 85.46 & 83.48 & 91.22 & 88.82 & 86.64 \\
Hy-MT3-30B-A3B-IQ2 & 91.44 & 90.08 & 83.37 & 81.32 & 90.81 & 87.74 & 85.74 \\
\bottomrule
\end{tabular}
}
\vspace{1mm}

\begin{minipage}{0.98\textwidth}
\tiny
\textit{Notes.} All scores are on a 0--100 scale. Except for IFMTB, scores are TransRubric scores averaged over the DeepSeek-V4.1-Flash and Qwen3.5-397B-A17B judges. All-33 covers all pairs among the 33 FLORES-200 languages, while Low-Res. covers low-resource languages paired with Chinese or English. Mand, WMTB, HMTB and IFMTB denote Mandarin$\Leftrightarrow$Minority translation, WildMTBench-202611, HardMTBench and IFMTBench, respectively. IFMTB is independently judged by GPT-OSS-120B.
\end{minipage}
\end{table*}

%% file: tables/languages.tex
\begin{table*}[!htbp]
\centering
\caption{Languages supported by Hy-MT3 (102 in total).}
\label{tab:languages}
\scriptsize
\renewcommand{\arraystretch}{1.12}
\setlength{\tabcolsep}{4pt}
\resizebox{\textwidth}{!}{
\begin{tabular}{lll@{\hspace{1.2em}}lll@{\hspace{1.2em}}lll}
\toprule
\textbf{Language} & \textbf{Code} & \textbf{Chinese Name} & \textbf{Language} & \textbf{Code} & \textbf{Chinese Name} & \textbf{Language} & \textbf{Code} & \textbf{Chinese Name} \\
\midrule
\multicolumn{9}{l}{\textbf{Supported since Hy-MT2 (33 languages)}} \\
\midrule
Chinese & \texttt{zh} & 中文 & Italian & \texttt{it} & 意大利语 & Khmer & \texttt{km} & 高棉语 \\
English & \texttt{en} & 英语 & German & \texttt{de} & 德语 & Burmese & \texttt{my} & 缅甸语 \\
French & \texttt{fr} & 法语 & Vietnamese & \texttt{vi} & 越南语 & Persian & \texttt{fa} & 波斯语 \\
Portuguese & \texttt{pt} & 葡萄牙语 & Malay & \texttt{ms} & 马来语 & Gujarati & \texttt{gu} & 古吉拉特语 \\
Spanish & \texttt{es} & 西班牙语 & Indonesian & \texttt{id} & 印尼语 & Urdu & \texttt{ur} & 乌尔都语 \\
Japanese & \texttt{ja} & 日语 & Filipino & \texttt{tl} & 菲律宾语 & Telugu & \texttt{te} & 泰卢固语 \\
Turkish & \texttt{tr} & 土耳其语 & Hindi & \texttt{hi} & 印地语 & Marathi & \texttt{mr} & 马拉地语 \\
Russian & \texttt{ru} & 俄语 & Traditional Chinese & \texttt{zh-TR} & 繁体中文 & Hebrew & \texttt{he} & 希伯来语 \\
Arabic & \texttt{ar} & 阿拉伯语 & Polish & \texttt{pl} & 波兰语 & Bengali & \texttt{bn} & 孟加拉语 \\
Korean & \texttt{ko} & 韩语 & Czech & \texttt{cs} & 捷克语 & Tamil & \texttt{ta} & 泰米尔语 \\
Thai & \texttt{th} & 泰语 & Dutch & \texttt{nl} & 荷兰语 & Ukrainian & \texttt{uk} & 乌克兰语 \\
\midrule
\multicolumn{9}{l}{\textbf{Chinese minority languages and dialects, supported since Hy-MT2 (5 languages)}} \\
\midrule
Tibetan & \texttt{bo} & 藏语 & Traditional Mongolian & \texttt{mn} & 传统蒙古语 & Cantonese & \texttt{yue} & 粤语 \\
Arabic Kazakh & \texttt{kk} & 阿拉伯字母哈萨克语 & Uyghur & \texttt{ug} & 维吾尔语 &  &  &  \\
\midrule
\multicolumn{9}{l}{\textbf{Newly supported in Hy-MT3 (64 languages)}} \\
\midrule
Greek & \texttt{el} & 希腊语 & Welsh & \texttt{cy} & 威尔士语 & Sinhala & \texttt{si} & 僧伽罗语 \\
Romanian & \texttt{ro} & 罗马尼亚语 & Catalan & \texttt{ca} & 加泰罗尼亚语 & Punjabi & \texttt{pa} & 旁遮普语 \\
Hungarian & \texttt{hu} & 匈牙利语 & Galician & \texttt{gl} & 加利西亚语 & Sindhi & \texttt{sd} & 信德语 \\
Bulgarian & \texttt{bg} & 保加利亚语 & Basque & \texttt{eu} & 巴斯克语 & Assamese & \texttt{as} & 阿萨姆语 \\
Serbian & \texttt{sr} & 塞尔维亚语 & Afrikaans & \texttt{af} & 南非荷兰语 & Odia & \texttt{or} & 奥里亚语 \\
Croatian & \texttt{hr} & 克罗地亚语 & Luxembourgish & \texttt{lb} & 卢森堡语 & Kannada & \texttt{kn} & 卡纳达语 \\
Slovak & \texttt{sk} & 斯洛伐克语 & Maltese & \texttt{mt} & 马耳他语 & Malayalam & \texttt{ml} & 马拉雅拉姆语 \\
Slovenian & \texttt{sl} & 斯洛文尼亚语 & Esperanto & \texttt{eo} & 世界语 & Sanskrit & \texttt{sa} & 梵语 \\
Albanian & \texttt{sq} & 阿尔巴尼亚语 & Yiddish & \texttt{yi} & 意第绪语 & Lao & \texttt{lo} & 老挝语 \\
Macedonian & \texttt{mk} & 马其顿语 & Armenian & \texttt{hy} & 亚美尼亚语 & Javanese & \texttt{jv} & 爪哇语 \\
Belarusian & \texttt{be} & 白俄罗斯语 & Georgian & \texttt{ka} & 格鲁吉亚语 & Sundanese & \texttt{su} & 巽他语 \\
Lithuanian & \texttt{lt} & 立陶宛语 & Azerbaijani & \texttt{az} & 阿塞拜疆语 & Cebuano & \texttt{ceb} & 宿务语 \\
Latvian & \texttt{lv} & 拉脱维亚语 & Uzbek & \texttt{uz} & 乌兹别克语 & Waray & \texttt{war} & 瓦瑞语 \\
Estonian & \texttt{et} & 爱沙尼亚语 & Kyrgyz & \texttt{ky} & 吉尔吉斯语 & Swahili & \texttt{sw} & 斯瓦希里语 \\
Finnish & \texttt{fi} & 芬兰语 & Turkmen & \texttt{tk} & 土库曼语 & Amharic & \texttt{am} & 阿姆哈拉语 \\
Swedish & \texttt{sv} & 瑞典语 & Tajik & \texttt{tg} & 塔吉克语 & Somali & \texttt{so} & 索马里语 \\
Danish & \texttt{da} & 丹麦语 & Tatar & \texttt{tt} & 鞑靼语 & Malagasy & \texttt{mg} & 马达加斯加语 \\
Norwegian & \texttt{nb} & 挪威语 & Bashkir & \texttt{ba} & 巴什基尔语 & Haitian Creole & \texttt{ht} & 海地克里奥尔语 \\
Norwegian Nynorsk & \texttt{nn} & 新挪威语 & Kurdish & \texttt{ku} & 库尔德语 & Cyrillic Kazakh & \texttt{kk-Cyrl} & 西里尔哈萨克语 \\
Icelandic & \texttt{is} & 冰岛语 & Central Kurdish & \texttt{ckb} & 中库尔德语 & Cyrillic Mongolian & \texttt{mn-Cyrl} & 西里尔蒙古语 \\
Irish & \texttt{ga} & 爱尔兰语 & Pashto & \texttt{ps} & 普什图语 &  &  &  \\
Scottish Gaelic & \texttt{gd} & 苏格兰盖尔语 & Nepali & \texttt{ne} & 尼泊尔语 &  &  &  \\
\bottomrule
\end{tabular}
}
\end{table*}